\documentclass[10pt,journal,compsoc]{IEEEtran}
\usepackage{times}
\usepackage{epsfig}
\usepackage{graphicx}
\usepackage{amsmath}
\usepackage{amssymb}
\usepackage{booktabs}
\usepackage{diagbox}
\usepackage{booktabs}
\usepackage[algo2e, vlined, ruled, linesnumbered]{algorithm2e}
\usepackage{comment}
\usepackage{times}
\usepackage{epsfig}
\usepackage{graphicx}
\usepackage{amsmath}
\usepackage{arydshln}
\usepackage{amssymb}
\usepackage{multirow}
\usepackage{xcolor}
\usepackage{enumitem}
\usepackage{array}
\usepackage{bbding}
\usepackage{subfigure}
\usepackage{url}
\usepackage{hyperref}
\hypersetup{hidelinks,
	colorlinks=true,
	allcolors=black,
	pdfstartview=Fit,
	breaklinks=true}

\DeclareMathOperator*{\argmax}{\arg\,max}

\newcommand{\vct}{\textrm{vec}}

\newcommand{\mbg}{\mathbb{G}}

\newcommand{\mbr}{\mathbb{R}}

\newcommand{\mbx}{\mathbb{X}}

\newcommand{\mca}{\mathcal{A}}
\newcommand{\mcc}{\mathcal{C}}
\newcommand{\mce}{\mathcal{E}}
\newcommand{\mcf}{\mathcal{F}}

\newcommand{\mcg}{\mathcal{G}}

\newcommand{\mcs}{\mathcal{S}}

\newcommand{\mcl}{\mathcal{L}}
\newcommand{\mcv}{\mathcal{V}}
\newcommand{\mci}{\mathcal{I}}

\newcommand{\mbfk}{\mathbf{K}}
\newcommand{\mbfx}{\mathbf{X}}
\newcommand{\mbfu}{\mathcal{U}}

\newcommand{\mbfm}{\mathbf{M}}

\newcommand{\our}{{\bf UPM}}

\ifCLASSOPTIONcompsoc
  \usepackage[nocompress]{cite}
\else
  \usepackage{cite}
\fi

\ifCLASSINFOpdf
\else
\fi

\begin{document}

\title{Learning Universe Model for Partial Matching Networks over Multiple Graphs}

\author{Zetian~Jiang,~
        Jiaxin~Lu,~
        Tianzhe~Wang,~
        and~Junchi~Yan,~\IEEEmembership{Senior~Member,~IEEE}% <-this % stops a space
\IEEEcompsocitemizethanks{\IEEEcompsocthanksitem Z. Jiang, J. Lu, T. Wang and J. Yan are with Department of Computer Science and Engineering, and MoE Key Lab of Artificial Intelligence, AI Institute, Shanghai Jiao Tong University, Shanghai, 200240, P.R. China.\protect\\
% note need leading \protect in front of \\ to get a newline within \thanks as
% \\ is fragile and will error, could use \hfil\break instead.
E-mail: \{maple\_jzt,lujiaxin,usedtobe,yanjunchi\}@sjtu.edu.cn\protect\\
Z. Jiang and J. Lu contribute equally to this work. \protect\\
Correspondence author: Junchi Yan.\protect
}}

% The paper headers
\markboth{Journal of \LaTeX\ Class Files,~Vol.~14, No.~8, October~2022}%
{Shell \MakeLowercase{\textit{et al.}}: Deep Reinforcement Learning of Graph Matching}
\IEEEtitleabstractindextext{
\begin{abstract}
We consider the general setting for partial matching of two or multiple graphs, in the sense that not necessarily all the nodes in one graph can find their correspondences in another graph and vice versa. We take a universe matching perspective to this ubiquitous problem, whereby each node is either matched into an anchor in a virtual universe graph or regarded as an outlier. Such a universe matching scheme enjoys a few important merits, which have not been adopted in existing learning-based graph matching (GM) literature. First, the subtle logic for inlier matching and outlier detection can be clearly modeled, which is otherwise less convenient to handle in the pairwise matching scheme. Second, it enables end-to-end learning especially for universe level affinity metric learning for inliers matching, and loss design for gathering outliers together. Third, the resulting matching model can easily handle new arriving graphs under online matching, or even the graphs coming from different categories of the training set. To our best knowledge, this is the first deep learning network that can cope with two-graph matching, multiple-graph matching, online matching, and mixture graph matching simultaneously. Extensive experimental results show the state-of-the-art performance of our method in these settings.
\end{abstract}

% Note that keywords are not normally used for peerreview papers.
\begin{IEEEkeywords}
Multiple Graph Matching, Partial Matching, Incremental Matching, Combinatorial Optimization 
\end{IEEEkeywords}}

% make the title area
\maketitle

%%%%%%%%%%%%%%%%%%%%%%%%%%%%%%%%%% Introduction %%%%%%%%%%%%%%%%%%%%%%%%%%%%%%%%%%%%%%%%%%%%%%%%%%%%%%%%%%%%%%%%
\section{Introduction}
\label{sec:introduction}
Finding consistent association among multiple objects is a fundamental task in pattern recognition and computer vision. The applications range from image registration~\cite{ShenTMI02}, structure from motion~\cite{bregler2000recovering, vijayanarasimhan2017sfm}, object tracking~\cite{nam2016learning, iqbal2017posetrack}, optical flow~\cite{baker2011database, sun2014quantitative, sun2018pwc}, stereo matching~\cite{luo2016efficient, chang2018pyramid}, and pose estimation~\cite{cao2017realtime, gasse2019exact}, etc.
In these tasks, not only the appearance but also the structure of the objects play an important role in establishing the correspondence. 

In particular, graph matching (GM) aims to find node correspondence between graph-structured data via second-order~\cite{Gold1996AGA, Cho2010ReweightedRW} or even high-order information~\cite{lee2011hyper}, which has shown its superiority on robustness against local noise. Specifically, Lawler’s quadratic assignment programming (QAP) is the most general form for two-graph matching:
\begin{equation}
\label{eq:gm_formulation}
\begin{aligned}
    &\mbfx = \argmax_\mbfx \vct(\mbfx)^{\top} \mbfk \vct(\mbfx) \\
    s.t. \hspace{5pt} \mbfx &\in \{0,1\}^{n_1 \times n_2}, \mbfx \mathbf{1}_{n_2} \leq \mathbf{1}_{n_1}, \mbfx^{\top} \mathbf{1}_{n_1} \leq \mathbf{1}_{n_2}
\end{aligned}
\end{equation}
where $\mbfx$ is a (partial) permutation matrix encoding node-to-node correspondence, and $\vct(\mbfx)$ is its column-vectorized version. $\mathbf{1}_n$ represents column vector of length $n$ whose elements all equal to $1$. $\mbfk \in \mathbb{R}^{n_1 n_2 \times n_1 n_2}$ is the affinity matrix (also called cost matrix in the equivalent minimization problem in literature). Its diagonal and off-diagonal elements store the node-to-node and edge-to-edge affinities, respectively. A popular extension to multiple graph matching (MGM)~\cite{yan2015consistency, ZhouICCV15, YanPAMI16, JiangPAMI21} can be written as:
\begin{equation}
\label{eq:mgm_formulation}
\begin{aligned}
    &\mathbb{X} = \argmax_{\mathbb{X} = \{\mbfx_{ij}\}_{i=1,j=1}^{N, N}} \sum \vct(\mbfx_{ij})^{\top} \mbfk_{ij} \vct(\mbfx_{ij}) \\
    s.t. \hspace{5pt} \mbfx_{ij} & \mathbf{1}_{n_j} \leq \mathbf{1}_{n_i}, \mbfx_{ij}^{\top} \mathbf{1}_{n_i} \leq \mathbf{1}_{n_j}, \mbfx_{ik} \mbfx_{kj} \leq \mbfx_{ij}, \forall i,j,k
\end{aligned}
\end{equation}
where cycle consistency~\cite{yan2015consistency,YanPAMI16} is often enforced. Both GM and MGM problems are in general NP-hard and many approximate algorithms have been proposed over the decades~\cite{YanICMR16,YanIJCAI20}.

\begin{figure}[tb!]
    \centering
    \includegraphics[width=0.5\textwidth]{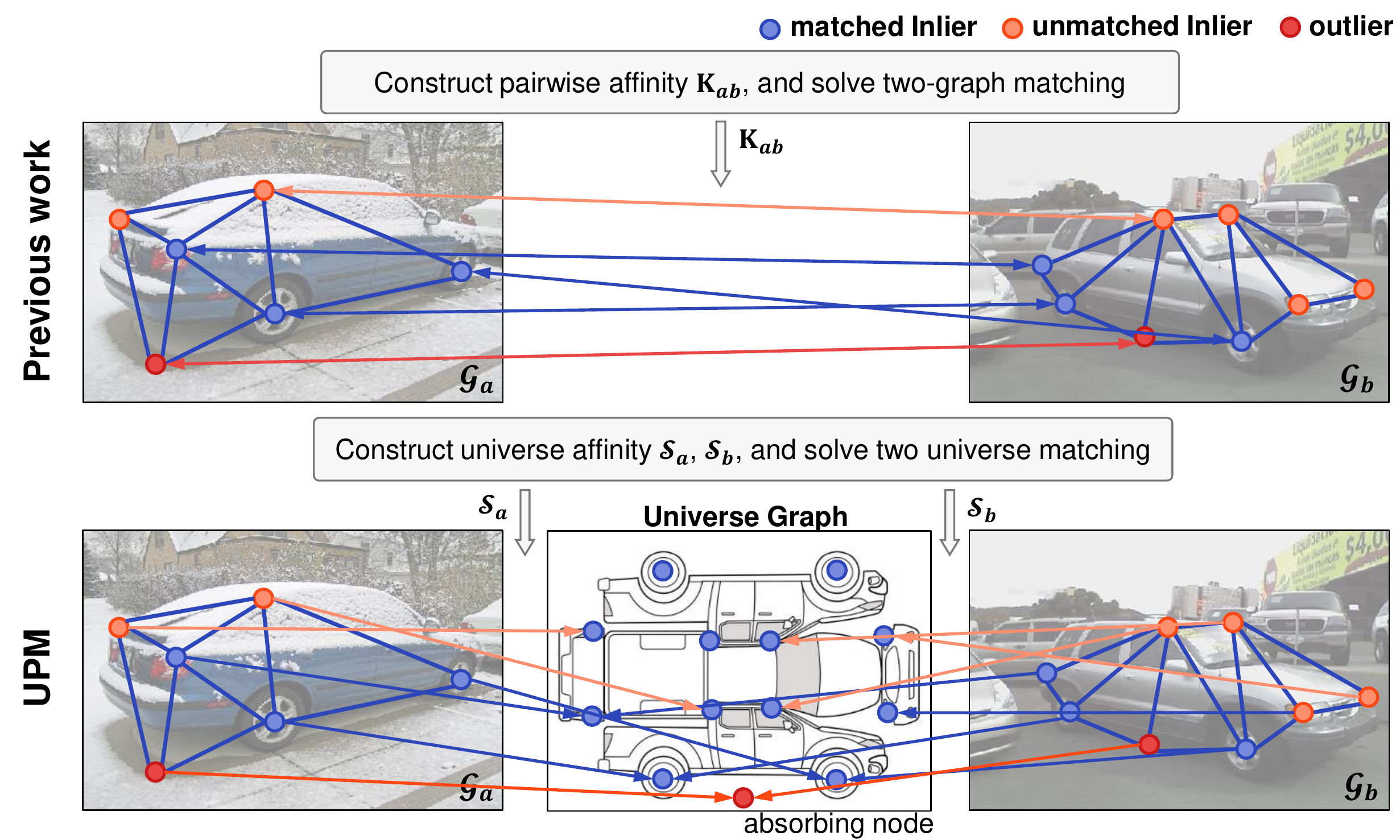}
    \caption{Teaser for our method \textbf{U}niverse \textbf{P}artial \textbf{M}atching (\our), a universe graph based partial matching solver. Compared with the pairwise matching model on the top, a universe graph with all semantic inliers and an absorbing node is introduced as a bridge for matching in \our. All the graphs are first matched to the universe graph and then pairwise matching is reconstructed via universe graph.}
    \label{fig:node_type}
  %  %\vspace{-10pt}
\end{figure}

Deep learning has shown its potential in pushing the frontier of GM for real-world image data~\cite{ZanfirCVPR18, WangICCV19, wang2020learning, Wang2019Neural, rolinek2020deep, yu2021deep}. Benefiting from the power of convolutional neural networks (CNNs) and graph neural networks (GNNs), expressive appearance and structure features are extracted from the source data. In these methods~\cite{ZanfirCVPR18, WangICCV19, Wang2019Neural}, it is often assumed that all the nodes can find their correspondences from the other graph, and vice versa, which means the matching between graphs is a bijective function. We refer to it as bijective matching in this paper.

However, the data obtained from real-world often breaks the assumption that holds for bijective matching:
\begin{itemize}[leftmargin=*]
    \item The graphs to be matched suffer `partial overlap' in most cases. It is hard to ensure all the graphs share nodes of the same kinds, e.g. in image matching tasks, key points could be occluded or missed due to rotation. 
    \item Although the nodes information is the input for graph matching problem, it is often manually labeled or detected by algorithm. Mistakes can not be avoided during such processing. For example, keypoint detection algorithm would generate coordinates for meaningless nodes, which are spurious and have no correspondence in graph matching settings.
    \item In some extreme cases, the input graphs are even of different classes (mixture graph matching and clustering~\cite{WangAAAI20}), which means they share no common nodes at all.
    % or they are given incrementally with time sequence (online MGM~\cite{YuECCV18}).
\end{itemize}

In this paper, we focus on a more general setting, so-called partial matching~\cite{chen2014near, ZhouICCV15}. Partial matching is an extension of bijective matching, where the correspondence between graphs is relaxed to a partial function. Compared with bijective matching, it is closer to practice and becomes a more challenging task. 

\begin{table}[tb!]
    \caption{Categorization of different nodes for matching. In a specific two-graph matching instance, node is divided to three types: matched inlier, unmatched inlier and outlier, based on whether it has semantic coherence and whether it has correspondence on another graph. Note both `unmatched inlier' and `outlier' are treated as spurious nodes in previous work, where their semantic meaning difference are ignored.}
    \label{tab:node_type}
    %\vspace{-5pt}
    \resizebox{0.49\textwidth}{!}{
    \begin{tabular}{|m{2.7cm}<{\centering}|p{1.2cm}<{\centering}|p{1.2cm}<{\centering}|p{1.2cm}<{\centering}|}
        \hline
                                            & Matched Inlier & Unmatched Inlier & Outlier       \\ \hline
        Semantic Coherence    & \Checkmark     & \Checkmark       & \XSolidBrush  \\ \hline
        Pairwise Matching      & \Checkmark     & \XSolidBrush     & \XSolidBrush  \\ \hline
    \end{tabular}}
  %  %\vspace{-10pt}
\end{table}

Partial matching has not been fully explored and one key lies in how to cope with spurious nodes and matchings \footnote{Spurious nodes refer to all the nodes that have no correspondence in matching instance, including both unmatched inlier and outlier.}. BBGM~\cite{rolinek2020deep} deals with partial matching via learning negative affinity to reduce the spurious matching pair, while DLGM~\cite{yu2021deep} also tries to predict consistent topology of graphs to alleviate the spurious map. All these previous works suffer from two drawbacks:
\begin{itemize}[leftmargin=*]
    \item They can only deal with two-graph matching, under which partial matching is an ill-posed problem and cannot be settled down perfectly. In the real-world, partial matching is more suitable to be formulated as multiple graphs matching, e.g. 3D reconstruction from multiple views and video tracing with several frames.
    \item They are often confused by the spurious nodes. There are two kinds of nodes having no correspondence in a two-graph matching instance: the nodes with semantic meaning but not shared in both graphs, and the meaningless nodes labeled or detected by mistakes. However, they are all treated as spurious nodes in previous work, which degrades the quality of the gradient and leads to poor performance.
\end{itemize}

To facilitate the discussion, we divide the nodes into three types: matched inlier, unmatched inlier, and outlier, where the last two are spurious nodes. Unmatched inlier denotes the node with semantic coherence but no correspondence, while outlier denotes the node labeled or detected by random mistake. The division  is shown in Table~\ref{tab:node_type}. More detailed mathematical definition is given in Sec.~\ref{subsec: node distinguishing}

\begin{figure}[tb!]
    \centering
    \includegraphics[width=0.5\textwidth]{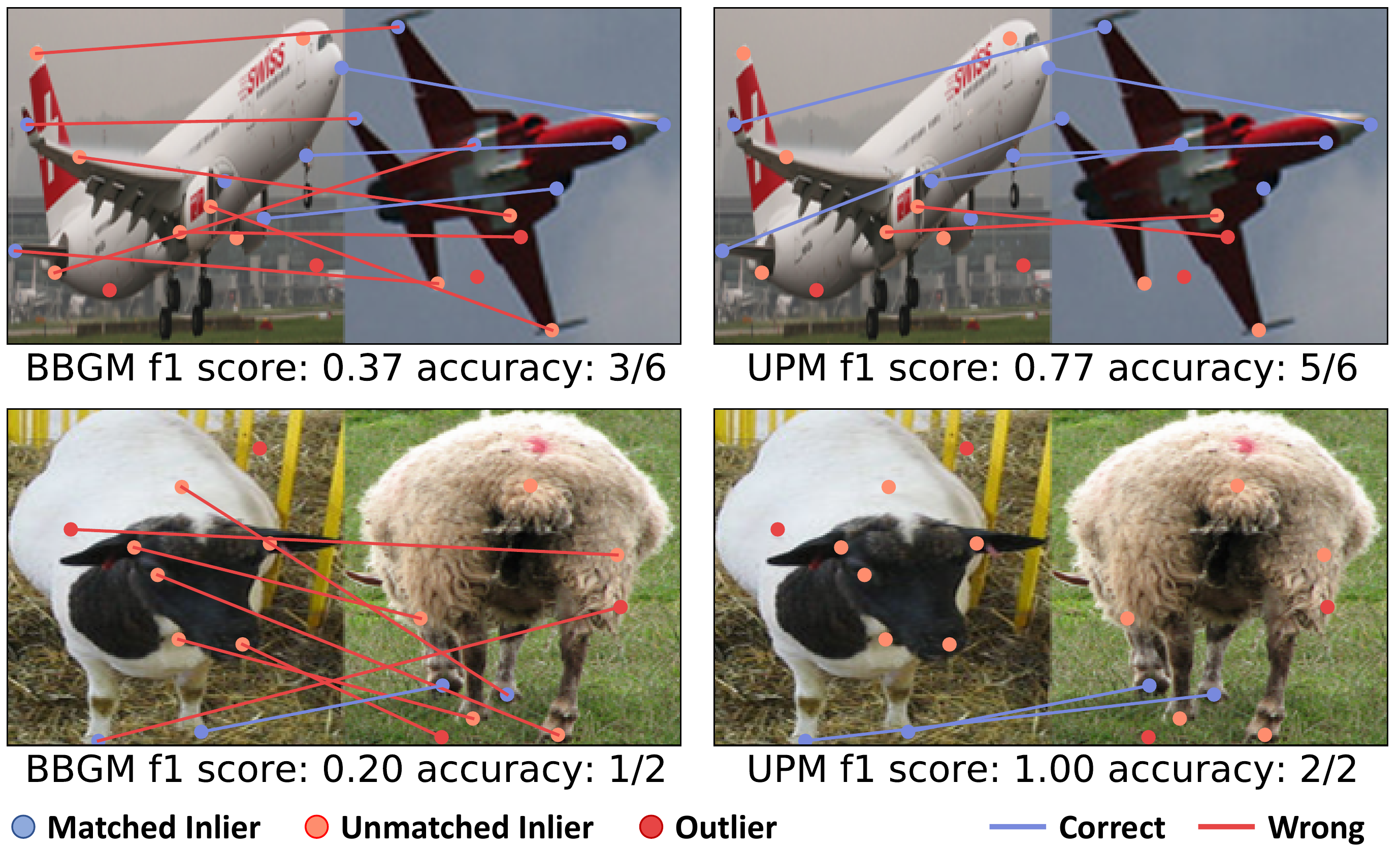}
    \caption{Two matching instances of BBGM and our method \our. Red lines denote the wrong correspondence, and the blue lines denote the correct correspondence. The matching results illustrate that our method is good at distinguishing spurious nodes. Compared with BBGM, the number of matching between spurious nodes is much less, and the overall matching quality is improved on both F1 score and accuracy.}
    \label{fig:show_instance}
  %  %\vspace{-10pt}
\end{figure}

To resolve the challenges above, we introduce the universe graph \cite{PachauriNIPS13,ZhouICCV15} into deep learning model and proposed a method, so-called \textbf{U}niverse \textbf{P}artial \textbf{M}atching (\our), as shown in Fig.~\ref{fig:node_type}. It can cope with both two-graph and multiple graph matching at the same time, and dedicated modules are designed to distinguish node type on a fine-grained level. More specifically, all the input graphs are matched to a universe graph instead of pairwise matching, which unifies the matching process of two-graph and multiple graphs, and cycle consistency is automatically satisfied. Meanwhile, the universe graph is also able to serve as a filter to distinguish unmatched inlier and outliers. Fig.~\ref{fig:node_type} shows that different nodes have different correspondence states on universe graph. The real-world matching instances are also shown in Fig. ~\ref{fig:show_instance} that our method \our \ take a great advantage in distinguishing spurious node types and reducing spurious matching prediction, compared with SOTA BBGM~\cite{rolinek2020deep}.

In a nut shell, the main contribution can be included as follows, and the source code of \our ~will be made public available at \href{https://github.com/Thinklab-SJTU/UniversePartialMatching}{https://github.com/Thinklab-SJTU/UniversePartialMatching}.
\begin{itemize}[leftmargin=*]
    \item We analyze the partial matching in detail. Under a multiple graph matching perspective, we propose the definition of unmatched inlier and outliers and reveal the limitation of pairwise graph matching on distinguishing them. We, therefore, introduce universe matching into deep learning for the first time.
    \item We design pipeline and loss for universe matching. More specifically, we propose the universe metric learning to distinguish the unmatched inliers and outlier-aware loss to filter out outliers. We also give detailed mathematical analysis on mechanism and gradient to illustrate how these modules work.
    \item Our proposed method \our \ outperforms SOTA of deep learning model and learning-free solver on main-stream datasets in the partial matching setting. It also has a strong ability on generalization and is robust for hyper-parameters.
    \item To our best knowledge, \our \ is also the first deep learning method that can deal with two-graph and multiple graph matching simultaneously. It can also be applied to extension tasks e.g. online MGM and mixture graph matching and clustering. Compared with previous work, \our \ achieves the best performance and notably accelerates the matching process on real-world benchmarks.
\end{itemize}

%%%%%%%%%%%%%%%%%%%%%%%%%%%%%%%%%% Related Work %%%%%%%%%%%%%%%%%%%%%%%%%%%%%%%%%%%%%%%%%%%%%%%%%%%%%%%%%%%%%%%%
\section{Related Work}
Graph matching has been a long-standing and active research area, from two-graph to multiple graph matching, and recently deep learning of graph matching, which we briefly review in this paper.
\subsection{Classical Two Graph Matching}

Graph matching refers to two-graph matching or pairwise graph matching, which aims to solve the Eq.~\ref{eq:gm_formulation} mentioned above. \cite{Gold1996AGA} solves graph matching via graduated assignment (GA) algorithm, which computes the partial derivative of the objective function with Taylor expansion to turn the GM problem into a linear assignment problem. RRWM~\cite{Cho2010ReweightedRW} constructs an association graph for matching graphs and applies the random walk algorithm on it with matching constraints. RRWHM~\cite{lee2011hyper} is the extended version to RRWM where the association graph is built with higher-order geometric information extracted from the graph. DS++, DS*~\cite{DymTOG17, bernard2018ds} design different ways to tightly relax GM to a convex problem, and then project the results back to the original solution space. LPMP~\cite{swoboda2017study} propose several additional Lagrangian relaxations of the graph matching problem and o leading solvers for this problem optimize the Lagrange decomposition duals with sub-gradient and dual ascent updates.

\subsection{Classical Multiple Graph Matching}
There are mainly two forms of methods to solve the Eq.~\ref{eq:mgm_formulation} of MGM. The first solves the problem in discrete space, which includes the composition-based method CAO, MGM-Floyd~\cite{YanPAMI16, JiangPAMI21} and the tree-structure-based approach MatchOpt~\cite{yan2015consistency}. These methods iteratively optimize the matching via finding the optimal matching chains over pairwise graph matching results, whereby the cycle consistency constraint is either automatically satisfied by the mechanism or encouraged in the objective function. 

Another line of works~\cite{HuangSGP13,chen2014near,wang2018multi} etc. try to relax the MGM problem into a continuous space and re-project the solution back to obtain a discrete matching result. The global consistency is maintained with a universe-graph-like structure and pursued via some continuous optimization methods, e.g. spectral clustering~\cite{PachauriNIPS13}, semidefinite programming~\cite{HuangSGP13} and further robust improvement~\cite{chen2014near} using alternating direction methods of multipliers (ADMM) that can address the partial multi-graph matching setting. Differing from the above approaches that separate the affinity-based two-graph matching and consistency-driven smoothing into two stages, some other works e.g. MatchALS~\cite{ZhouICCV15, SwobodaCVPR19} also explore the joint optimization regarding consistency and affinity by relaxation in continuous space. 

Some extension settings of MGM are also explored as incremental multiple graph matching~\cite{YuECCV18, JiangPAMI21} or multiple graph matching and clustering~\cite{WangAAAI20, wang2020graduated}. Incremental multiple graph matching solver~\cite{YuECCV18, JiangPAMI21} aims to design an efficient algorithm for the setting where graphs are given one by one in sequence. Meanwhile, \cite{WangAAAI20, wang2020graduated} propose the method to deal with graphs of mixed class as input, where matching and clustering are often iteratively performed to improve the accuracy of both sides.

\subsection{Deep Learning of Graph Matching}
Deep learning has recently been applied to graph matching on images \cite{ZanfirCVPR18}, whereby CNN is used to extract node features from images followed with spectral matching and is learned using a regression-like node correspondence supervision. This work is improved via introducing GNN to encode structural~\cite{WangICCV19, WangPAMI20} or geometric~\cite{ZhangICCV19} information, with a combinatorial loss based on cross-entropy loss, and Sinkhorn network~\cite{AdamsArxiv11} as a differential matching solver. The work~\cite{YuICLR20} extends PCA (Permutation loss and Cross-graph Affinity GM)~\cite{WangICCV19} by edge embedding and Hungarian-based~\cite{KuhnNavalResearch55} attention mechanism to stabilize end-to-end training. BBGM (Blackbox Deep Graph Matching)~\cite{rolinek2020deep} proposes a better front-end feature extraction backbone with Spline Convolution~\cite{FeyCVPR18}, and the gradient is backpropagated by fitting linear gradient of the discrete graph matching solver~\cite{poganvcic2019differentiation}. NGM (Neural Graph Matching Nets)~\cite{Wang2019Neural, wang2020learning} proposes to address the most general Lawler's QAP form, based on the novel feature extractors e.g.~\cite{RolinekECCV20} with the proposed learnable graph matching solver. DLGM (Deep Latent Graph Matching)~\cite{yu2021deep}, based on BBGM, predicts consistent topology of graphs utilizing both deterministic and generative models to improve the matching quality.

\subsection{Partial Matching and Maximum Common Subgraph}
% There are two tasks similar to partial matching: Maximum common subgraph~\cite{yang2014weighted, ying2020neural, chen2020can, bai2021glsearch} and node labeling~\cite{sole2011models, sole2013graduated}. 

Maximum common subgraph (MCS) is similar to two-graph partial matching since we can see the nodes in the common subgraph as inliers and others as outliers. Learning-based methods are also introduced for MCS. However, it relies on graph isomorphism and aims to find the same common subgraph for inputs, while in partial matching, the geometric structure is often broken with massive outliers. \cite{ying2020neural} decomposes target graph by extracting k-hop neighborhood around and applying GNN to encode the structure of subgraph for matching. \cite{bai2021glsearch} introduces Q-Network (DQN) to replace the node selection heuristics required in state-of-the-art MCS solvers, which is more accurate and efficient.

%%%%%%%%%%%%%%%%%%%%%%%%%%%%%%%%%% Method %%%%%%%%%%%%%%%%%%%%%%%%%%%%%%%%%%%%%%%%%%%%%%%%%%%%%%%%%%%%%%%%
\section{Methodology}
\begin{figure*}[tb!]
    \centering
    \includegraphics[width=\textwidth]{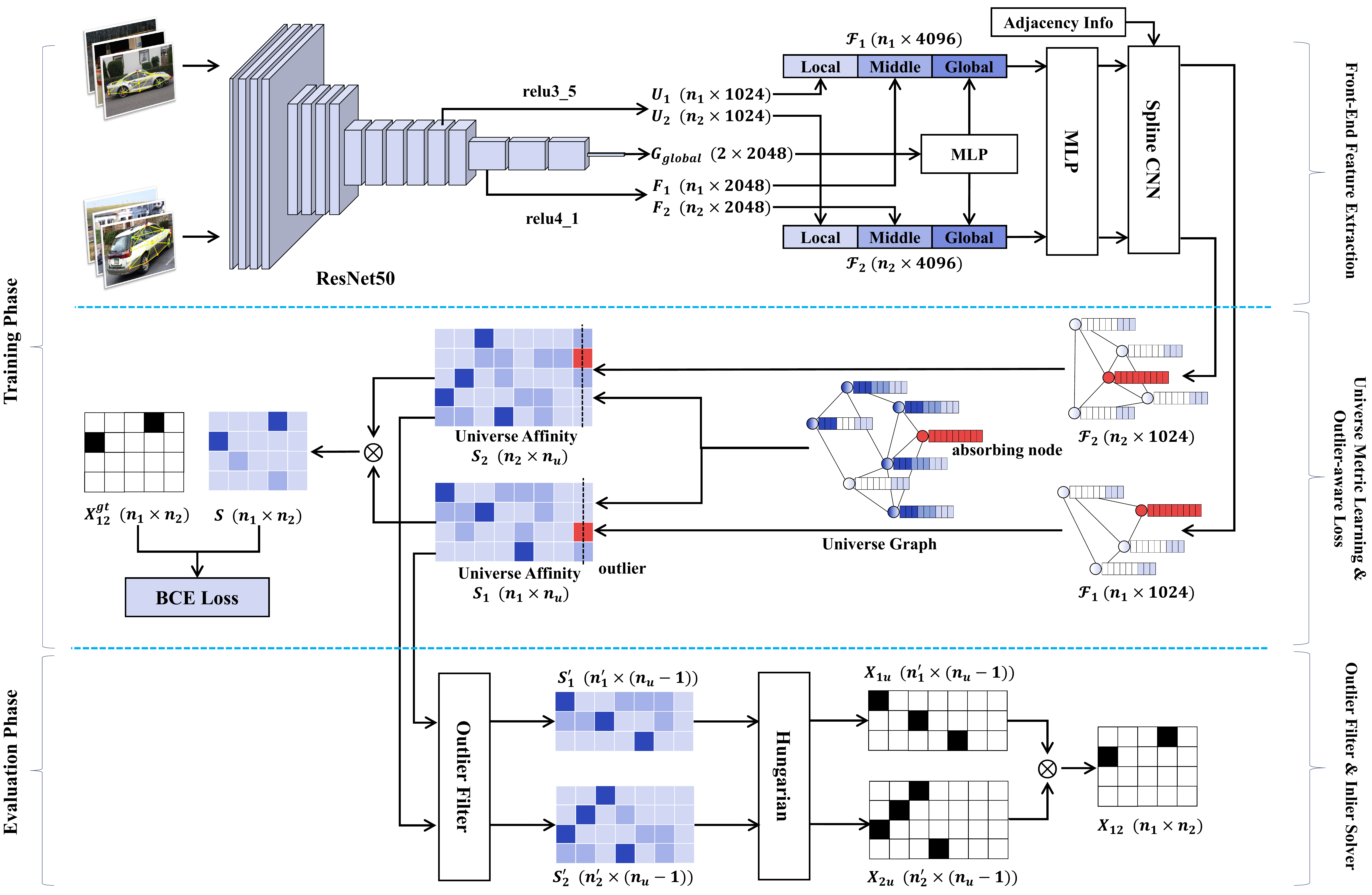}
    \caption{Overall pipeline for \our. The pipeline mainly consists of three parts: front-end feature extraction, universe metric learning, and universe matching solver. In front-end feature extraction, we extract local, middle, global features for each node with ResNet50~\cite{he2016deep} backbone and input them to Spline CNN~\cite{FeyCVPR18} to obtain node feature. In the universe metric Learning part, we build universe affinity between input graphs and universe graph respectively and train the universe graph with an outlier-aware loss, which will be discussed in detail in Sec.~\ref{subsec:pipeline} and Sec.~\ref{subsec:loss_design}. The red nodes in universe is absorbing node and red element in matrix illustrates that it is an outlier. In the evaluation phase, outlier filter and inlier solver are adopted to obtain a discrete solution. Outlier filter removes the last column and the rows of outliers. Inlier solver applies the Hungarian algorithm~\cite{KuhnNavalResearch55} to obtain universe matching results. The pairwise is reconstructed from universe matching.}
    \label{fig:pipeline}
    % %\vspace{-10pt}
\end{figure*}

In this section, we first present the motivation for introducing universe graph into deep learning in Sec.~\ref{subsec:universe_matching}. Then we introduce the design of the modules for universe matching with mathematical analysis. Universe metric learning is discussed in Sec.~\ref{subsec:pipeline} and outlier-aware loss is introduced in Sec.~\ref{subsec:loss_design}. Finally, we discuss the limitations of our model \our \  in Sec~\ref{subsec:extension&limitation}. The overall pipeline is shown in Fig.~\ref{fig:pipeline}, and the detailed algorithm is shown in Alg.~\ref{alg:main}.

\begin{table}%[!tb] 
\caption{Main notations and description used in this paper.}
\centering
    \begin{tabular}{l m{160pt}}
        \toprule[1pt]
        \textbf{Notations} & \textbf{Descriptions} \\
        \hline
        $\mbg$ & $\mbg$ is a set of $N$ graphs $\mbg = \{\mcg_{1} \dots \mcg_{N}\}$ \\
        \hline
        $\mcg$ & A graph instance $\mcg$ consists of vertex set $\mcv$ and edge set $\mce$. \\
        \hline
        $v$ & $v$ denotes the node on the graph. \\
        \hline
        $\mbfu$, $\mbfu_i$ & $\mbfu$ is the universe graph which consists of $n$ anchor points features $\mbfu = \{\mcf^{1}_{u} \dots \mcf^{(n)}_{u}\}$. $\mbfu_i$ is the sub-universe graph of $i$-th class. \\
        \hline
        $\mcf_a$,$\mcf_u$ & $\mcf$ denotes the node feature of $\mcg_a$ while $\mcf_u$ denotes the anchor features of universe graph. \\
        \hline
        $\mbx$ & $\mbx$ denotes all the pairwise matching results in graph set $\mbg$: $\mbx = \{\mbfx_{ij}\}_{i,j=1}^{N, N}$ \\
        \hline
        $\mbfx_{ij}, \mbfx_{iu}$ & $\mbfx_{ij}$ denotes pairwise matching between $\mcg_{i}$ and $\mcg_{j}$. $\mbfx_{iu}$ denotes universe matching from $\mcg_{i}$ to universe graph. \\
        \hline
        $\hat{\mbx}$, $\hat{\mbfx}_{ij}$, $\hat{\mbfx}_{iu}$ & The matching with hat denotes the prediction made by model. \\
        \hline
        $\mbfk_{ab}$ & $\mbfk_{ab} \in \mathbb{R}^{n_a n_b \times n_a n_b}$ denotes the affinity matrix between $\mcg_{a}$ and $\mcg_b$. Its diagonal and off-diagonal elements store the node-to-node and edge-to-edge affinities, respectively. \\
        \hline
        $\mcs_{ab}$, $\mcs_{a}$ & $\mcs_{ab} \mathbb{R}^{n_a \times n_b}$ denotes the node affinity between $\mcg_{a}$ and $\mcg_b$. $\mcs_{a}$ denotes the universe affinity from $\mcg_{a}$ to universe graph.\\
        \hline
        $p_{i}$ & $p_{i}$ denotes the matching likelihood distribution from node $v_i$ to universe graph $\mbfu$. \\
        \hline
        $p_{ij}$ & $p_{ij}$ is the matching probability of $v_i$ and $v_j$. \\
        \hline
        $\mcl$, $\mcl_{ij}$ & $\mcl$ denotes the overall loss and $\mcl_{ij}$ denotes the loss between matching pair $\mcg_i$ and $\mcg_j$. \\
        % \hline
        \bottomrule[1pt]   
    \end{tabular}
    \label{tbl-notation}
\end{table}

%%%%%%%%%%%%%%%%%%%%%%%%%%%%%%%%%% Universe Matching %%%%%%%%%%%%%%%%%%%%%%%%%%%%%%%%%%%%%%%%%%%%%%%%%%%%%%%%%%%%%%%%

\subsection{Motivation for the Universe Graph Perspective}
\label{subsec:universe_matching}
The idea of the universe graph has once been discussed in \cite{PachauriNIPS13,ZhouICCV15}. It is a virtual graph where all the nodes of input graphs can find their correspondence on universe graph. An absorbing node is added to the universe graph, which is matched to all the outliers. We introduce the universe graph into deep learning model for two reasons: 1) universe graph can aggregate the information over all the graphs in the training dataset, thus providing a comprehensive view for node type distinguishing; 2) universe graph unifies the matching procedure for both two-graph and multiple graph, which allows a unified model for all the matching tasks in graph matching.

\subsubsection{Global view for node distinguishing}\label{subsec: node distinguishing}
We begin with the node types applied in pairwise matching-based models~\cite{WangICCV19, WangPAMI20, rolinek2020deep, yu2021deep}. In a two-graph matching instance, whether nodes are spurious or not is decide by the pairwise graph correspondence (\texttt{PGC}). \texttt{PGC} denotes whether a node has correspondence in another graph. Given a pair of graphs $(\mcg_a, \mcg_b)$ with ground truth matching $\mbfx_{ab}^{gt}$, \texttt{PGC} of $v_i \in \mcg_{a}$ is defined as:
\begin{equation} \label{eq:mgm_correspondence}
    \texttt{PGC}(v_i | \mcg_a, \mcg_b, \mbfx_{ab}^{gt}) = 
    \left\{
        \begin{array}{cc}
            1, & \sum_{v_j \in \mcg_b} \mbfx_{ab}^{gt}(v_i, v_j) = 1 \\
            0, & \sum_{v_j \in \mcg_b} \mbfx_{ab}^{gt}(v_i, v_j) = 0 
        \end{array}
    \right.
\end{equation}
This kind of node division is rough and ignores the semantic differences of spurious nodes. 

In previous works~\cite{rolinek2020deep, yu2021deep}, the matched nodes are clustered together while all the spurious nodes are excluded from each other in feature space. However, the type of nodes changes from case to case, e.g. the matched node in one case can turn to the spurious nodes in another matching instance. Therefore, the gradient directions from two scenarios can be totally different or even conflicted. 
Meanwhile, outliers are always spurious and excluded from each other. As the consequence, they are uniformly scattered in the feature space after training and are easily confused with other key points. In fact, outliers need to be clustered and handled together to obtain a better solution. 
All in all, nodes of different types require different strategies, and thus we propose a fine-grained level division for spurious nodes. 

As mentioned in Table~\ref{tab:node_type}, matched nodes are called matched inlier in this paper, while spurious nodes are further divided into unmatched inlier and outlier via semantic coherence. 
We define semantic coherence based on the assumption that a node with specific semantic coherence can reoccur, while a node obtained by random mistakes is unique. 
That is to say, a node with semantic coherence is bound to have correspondence with other nodes, as long as the dataset is large enough. Therefore,  multiple graph correspondence (\texttt{MGC}) is proposed to define whether a node has semantic coherence as follows:
\begin{equation} \label{eq:mgm_correspondence}
    \texttt{MGC}(v_i | \mcg_a, \mathbb{G}, \mathbb{X}^{gt}) = 
    \left\{
        \begin{array}{cc}
            1, & \sum_{\mcg_b \in \mathbb{G}} \sum_{v_j \in \mcg_b} \mbfx_{ab}^{gt}(v_i, v_j) > 0 \\
            0, & \sum_{\mcg_b \in \mathbb{G}} \sum_{v_j \in \mcg_b} \mbfx_{ab}^{gt}(v_i, v_j) = 0 
        \end{array}
    \right.
\end{equation}
where $\mathbb{G}$ denotes all the graphs in training data and $\mathbb{X}^{gt}$ denotes all the ground truth matching results.

Note that \texttt{MGC} is defined under the view of multiple graph with all the graphs $\mathbb{G}$ and ground truth matching results $\mathbb{X}^{gt}$. Therefore, two-graph base models can not resolve it well. That's why we introduce the universe graph to aggregate the global information of all graphs in training data. As shown in Fig.~\ref{fig:node_type}, different node type has distinctive performance, which illustrates that node type is easy to be distinguished via a universe graph. Furthermore, the universe graph offers a template for the input graph. Each node is encouraged to match to an anchor node on the universe. It avoids the conflict gradient from the different matching instances, which occurs in previous works.

\subsubsection{Unified model for graph matching}

There are several tasks in graph matching, e.g. two-graph matching, multiple graph matching, online matching, and mixture graph matching. All of them can be solved by our proposed method \our, due to the merits of universe graph.

The key difference between two-graph matching and multiple graph matching is the cycle consistency~\cite{ZhouICCV15, YanPAMI16}, e.g. the composition of matches along a loop of graphs should be identical. In our method \our, universe graph serves as a center where all the input graphs are matched to universe graph first and then pairwise graph are reconstructed from universe matching results. Cycle consistency is automatically satisfied with the centralized structure. Therefore, two-graph matching and multiple graph matching are unified via universe graph.

Online MGM, also called incremental MGM, is recently proposed by \cite{YuECCV18}. Rather than being obtained at one time, graphs are often collected over time in practice, e.g. photos taken by street-view vehicles, video from the surveillance camera, newly discovered protein v.s. existing protein. For this setting, the naive strategy by matching all the old graphs with new coming graphs is inefficient. Moreover, it also needs post-process if cycle consistency is required. As for the universe graph, only one forward between the universe graph and the new coming one is enough to obtain the overall matching results, which reduces the time cost from linear magnitude to constant magnitude. Meanwhile, it is also space-efficient. Since the pairwise matching can be reconstructed by universe matching, we only need to keep the latter. This reduces the space storage from square to linear order.

On another line of the problem, most matching tasks assume that all the graphs belong to the same category and contain common inliers, which is often too ideal in practice. Mixture graph matching and clustering~\cite{WangAAAI20} is proposed where multiple graphs of different categories are given as input. It can be seen as a kind of partial matching where the key points from different clusters do not match with each other, as unmatched inliers. Universe graphs can distinguish node types and thus can reject the pair of different categories. Therefore, \our \ is also able to cope with mixture graph matching and clustering.

In general, universe matching is robust and efficient for all kinds of tasks. All kinds of matching tasks can be unified under universe graph and thus \our \ serves as a unified model for graph matching.

%%%%%%%%%%%%%%%%%%%%%%%%%%%%%%%%%% Pipeline %%%%%%%%%%%%%%%%%%%%%%%%%%%%%%%%%%%%%%%%%%%%%%%%%%%%%%%%%%%%%%%%
\subsection{Metric Learning for Universe Graph}
\label{subsec:pipeline}

In this section, we first study how to construct universe affinity, and then further explore the structure of universe graph over multiple categories.

%%%%%%%%%%% Front-end Feature Extractor %%%%%%%%%%%%%%%%%%%
% \subsubsection{Front-end feature extractor}
% We apply the extractor in \cite{rolinek2020deep} with a few adjustments to accommodate  universe matching. The process is introduced in below:
% \begin{itemize}[leftmargin=*]
%     \item Compute the outputs of $\texttt{relu3\_5}$, $\texttt{relu4\_1}$ and the final layer of the RestNet50~\cite{he2016deep} network pre-trained on ImageNet~\cite{krizhevsky2012imagenet} as local, middle and global feature, correspondingly. Concatenate them to obtain the node feature:
%     \begin{equation}
%         \mcf = \texttt{CONCAT}(U,F,G)
%     \end{equation}
%     \item Pass node feature and graph adjacency as the input for geometric feature refinement part. The graph adjacency $\mca$ is created by Delaunay triangulation~\cite{delaunay1934sphere} of the keypoint locations. We apply SplineConv~\cite{FeyCVPR18} to further encode class embedding, higher order information and geometric structure of the whole graph into the node-wise feature.
%     \begin{equation}
%         \mcf = \texttt{SplineCNN}(\mcf, \mca)
%     \end{equation}
% \end{itemize}
% The pipeline of feature extractor is shown in Fig. \ref{fig:pipeline} and it shares weight for all matching objects.

%%%%%%%%%%% Universe metric learning %%%%%%%%%%%%%%%%%%%
\subsubsection{Construction of universe affinity}

\begin{figure*}
    \centering
    \includegraphics[width=0.8\textwidth]{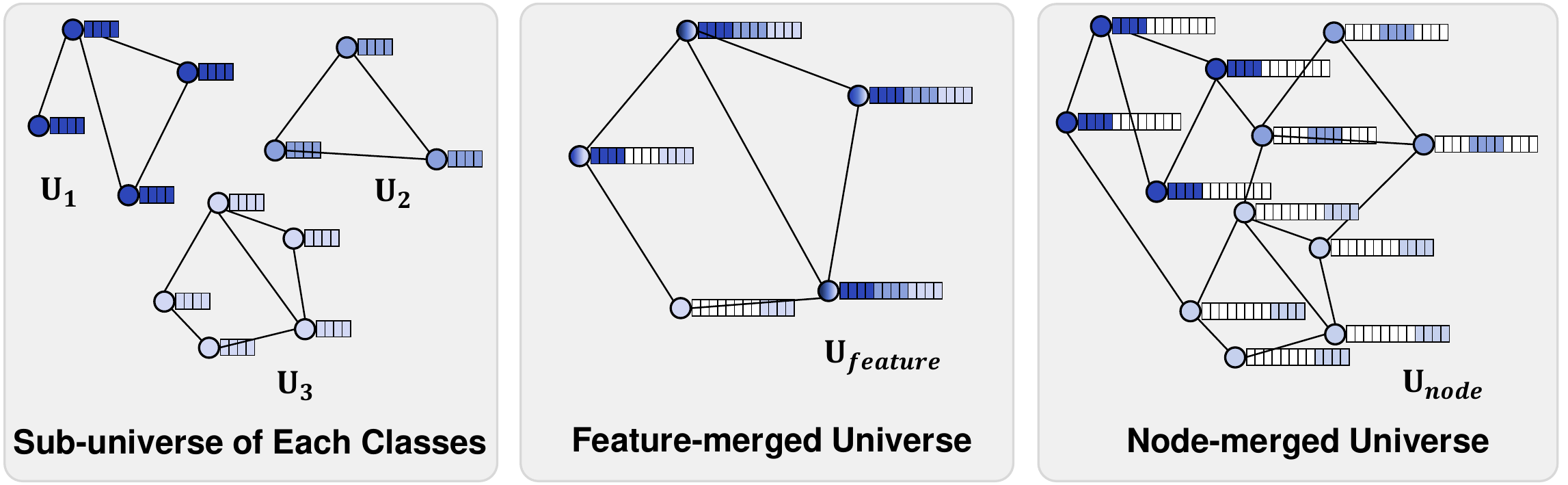}
    \caption{Illustration of two ways of universe graph construction. There are three sub-universe graphs with nodes and features of different colors. The second and the third graph show how universe graph merges these sub-universe graphs. The blank rectangles in universe graph represent the wasted channel of the universe anchor feature $\mcf_i$. The color of nodes in the feature-merged universe graph denotes that several nodes of sub-universe graphs are aggregated together. Note we aim to show the concatenation on node features. Edge features are ignored and the adjacency is meaningless.}
    \label{fig:universe_graph}
\end{figure*}

According to affinity matrix factorization proposed by \cite{ZhouPAMI16}, previous works \cite{ZanfirCVPR18, Wang2019Neural, rolinek2020deep} build the pairwise affinity $\mcs$ by:
\begin{equation}
    \mcs_{ab} = \mcf_{a} \mathbf{\Lambda} \mcf_{b}^{\top}
\end{equation}
where $\mcf_{a} \in \mbr^{n_a \times d},\mcf_{b}\in \mbr^{n_b \times d}$ denote the features of graph $\mcg_a$, $\mcg_b$ accordingly, and $\mathbf{\Lambda}$ is also a learnable weight of this affinity function. Such an affinity construction focuses on the similarity between the pair of graphs and is more suitable for the bijective matching setting. As for partial matching, massive outliers will destroy the geometric similarity between graphs, resulting in a large gap in feature extraction. The similarity assumptions of matching graphs followed in bijective matching will be severely broken, thus degrading the quality of the pairwise affinity. 

In this paper, we propose a new metric learning method, called `universe affinity', to overcome the challenges brought by partial matching. Specifically, we fabricate a universe graph $\mbfu$ with universe anchor features $\{\mcf_{u}^{(1)} \dots \mcf_{u}^{(n_u)} \}$ and try to learn the affinity $S_a$ between a single graph and the universe graph.
\begin{equation}
    \mcs_a = \mcf_{a} \mathbf{\Lambda}_{u} \mcf_{u}^{\top}
\end{equation}
where $\mathbf{\Lambda}_{u}$ and $\mcf_{u}$ share weight for all matching instance. Let $\mbfm=\mathbf{\Lambda}_{u} \mcf_{u}^{\top}$ be a learnable weight, it turns out:
\begin{equation}
    \mcs_a = \mcf_{a} \mbfm
\end{equation}
This is equivalent to performing a linear transformation on the features. We also add \textit{batch norm} and \textit{relu} to increase the non-linearity in metric learning, which can extract universe affinity of higher quality.

Compared with node features extracted from a single source graph, universe anchor features $\mcf_u$ aggregate the information of all the graphs in training data and thus is more robust against the node and structure noise. Therefore, the universe affinity constructed by anchor feature is more reliable. Universe anchor features separate the latent feature space into different parts and encourage the node features within the same region to be matched. 

Meanwhile, universe affinity does not lose information compared to pairwise affinity, because pairwise affinity could be easily reconstructed by universe affinity as follows:
\begin{equation} \label{eq:ua2pa}
    \mcs_{ab} = \mcf_{a} \mathbf{\Lambda} \mcf_{b}^{\top} = \mcf_{a} \mbfm \mbfm^{\top} \mcf_{b}^{\top} = \mcs_{a} \mcs_{b}^{\top}
\end{equation}

Although universe affinity could be applied to both node and edge affinity, we give up the edge affinity learning for three reasons. 1) Due to massive outliers in partial matching, the structural similarity of two graphs is badly broken. Edge affinity learning hardly improves or even harms the matching results. 2) Node features have already included geometric information that is carried by edge affinity for applying Spline CNN refinement module. 3) The number of edges on the universe graph is extensive, especially for the node-merged universe graph. The training phase of edge affinity is unstable and converges very slowly. Therefore, we only focus on node affinity for matching in this paper.

\subsubsection{Structure of universe graph}

Graph matching aims to explore the graph similarity between objects, which is able to cope with matching instances from different classes, e.g. the tasks of matching cars and matching boats can be solved with one model. However, each class would generate one universe graph and correspondent universe anchor features of their own. It calls for further exploration of how these universe graphs from different classes fuse together.

To facilitate the discussion, we called the universe graph from each class as sub-universe graph $\mbfu_i$. Its universe anchor features are represented as $\{\mcf_{i}^{(1)}, \dots, \mcf_{i}^{(n_i)}\}$, where $\mcf_{i}^{(j)} \in \mbr^{d}$ is node feature of $v_{j} \in \mbfu_{i}$ and $n_i$ is the node number of $\mbfu_{i}$. The universe graph $\mbfu$, which refers to the fusion graph of all sub-universe graphs, is represented as the same way: $\{\mcf_{u}^{(1)}, \dots, \mcf_{u}^{(n)} \}$. Note we only care about the node features of universe graph and the edge feature as well as adjacency is ignored during the discussion.

In general, we propose two hypotheses on the structure of universe graph cross all classes: feature-merged universe graph and node-merged universe graph, as shown in Fig.~\ref{fig:universe_graph}.

\textbf{Feature-merged Universe Graph} \hspace{3pt} 
Let $n$ be the maximum size of the sub-universe graphs $n = \max n_i$ where universe anchor feature $\mcf_u^{(i)}$ aggregates all the $i$-th features in each sub-universe graph:
\begin{equation}
    \mcf_u^{(i)} = \texttt{CONCAT}(\mcf_1^{(i)}, \dots, \mcf_m^{(i)})
\end{equation}
The feature number $n$ of $\mbfu$ stays the same with the increasing classes number. The total feature dim also has an upper bound to reach its maximum expression capability (limited by the front-end model). Therefore, this kind of universe graph construction method is more efficient in both time and memory. It is more suitable for situations where there are many graph categories or the relationship between categories is complex. 

\textbf{Node-merged Universe Graph} \hspace{3pt} 
Let $n$ be the sum of node number of sub-universe graphs $n = \sum n_i$ where overall universe graph $\mbfu$ aggregates all features of sub-universe graphs.
\begin{equation}
    \mbfu = \bigcup_i \mbfu_i = \bigcup_i \{\mcf_{i}^{(1)}, \dots, \mcf_{i}^{(n_i)}\}
\end{equation}
This type of universe graph construction method is more sensitive to graphs of different classes. It can reject two graphs from different categories and is more suitable for a mixture graph setting. However, the drawback is the size of universe graph, which increases linearly with classes number $m$. Moreover, due to all the features being extracted from a share-weighted front-end model, the feature dim should be consistent with input graph features, which means many channels are wasted. As shown in the right figure in Fig.~\ref{fig:universe_graph}, the feature dim is also 9, and 6 of them are wasted for each node on node-merged universe graph.

In practice, we can shift these two constructions by changing the size of universe graph and the associated training protocol. For feature-merged universe graph, we can estimate the maximum feature dimension of all the classes and sample the training pair of the same category. For node-merged universe graph, we estimate the sum of feature numbers over the classes, and half of the training pair is of the different categories. The feature channel assignment is adaptively learned with the model itself and we only set the total feature dim as a constant. We conduct experiments to further prove the construction of the universe graph conforms to our conjectures in both settings.

%%%%%%%%%%% Loss Design for Universe Matching %%%%%%%%%%%%%%%%%%%
\subsection{Loss Design for Outlier Clustering}
\label{subsec:loss_design}

Intuitively, outliers should be clustered and processed together to obtain a better solution. However, the major challenges lie in that outliers are never matched with each other, which lacks guidance to gather them. Therefore, outliers are excluded from each other and randomly scattered in the feature space in the previous work. 

In our method \our, we offer an absorbing node on universe graph as an anchor for outlier and design an outlier-aware loss to encourage outlier to get close to the anchor. In this section, we will begin at the vanilla loss analysis to show the rationality of our loss design, and then modify it to an outlier-aware loss.

\subsubsection{Vanilla loss design}
Learning-based GM problem can be viewed as a process of maximizing the likelihood, as formulated in \cite{yu2021deep}. Given a batch of training samples $\{\mbfx_{ab}^{(l)}, G_{a}^{(l)}, G_{b}^{(l)} \}$ with $l = 1, 2, ...,N$, the objective of learning-based GM aims to maximize the likelihood:
\begin{equation}
    \argmax_{\theta} \prod_l P_{\theta}(\hat{\mbfx}_{ab} = \mbfx^{(l)}_{ab} | \mcg_{a}^{(l)}, \mcg_{b}^{(l)}),
\end{equation}
where $\hat{\mbfx}_{ab}$ is the matching results predicted by the deep learning model. If we relax permutation constraints on results $\hat{\mbfx}_{ab}$, the probability can be factorized to node-level:
\begin{equation}
    P_{\theta}(\hat{\mbfx}_{ab} = \mbfx_{ab} | \mcg_{a}, \mcg_{b}) = \prod_{ij} P_\theta(\hat{x}_{ij} = x_{ij} | \mcg_{a}, \mcg_{b}).
\end{equation}
$\hat{x}_{ij}$ and $x_{ij}$ denotes the entry of $\hat{\mbfx}_{ab}$ and $\mbfx_{ab}$ accordingly, which represents the matching probability between $v_i$ and $v_j$. Note $\mbfx_{ab}$ is the ground truth label and satisfies the matching constraints. The prediction $\hat{\mbfx}_{ab}$ will be guided by $\mbfx_{ab}$, and thus satisfies matching constraints too. 

Let $p^{(k)}_{ij} = P_{\theta}(\hat{x}_{ij}^{(k)} = 1)$ denote the matching probability between $v_i$ and $v_j$. We can further rewrite the optimization target as follows:
\begin{equation}
%\begin{small}
\begin{aligned}
    & \argmax_{\theta} \prod_k P_{\theta}(\mbfx^{(k)} | \mcg_{a}^{(k)}, \mcg_{b}^{(k)}) \\
     = &\argmax_{\theta} \sum_k \sum_{ij} \log_{\theta}(\hat{x}_{ij}^{(k)} = x^{(k)}_{ij}) \\
     = &\argmax_{\theta} \sum_k \sum_{ij} x^{(k)}_{ij} \log(p^{(k)}_{ij}) + (1-x^{(k)}_{ij}) (1-\log(p^{(k)}_{ij})),
\end{aligned}
%\end{small}
\end{equation}
which has the same form as binary cross entropy. Therefore we apply BCE loss into learning-based GM problem as in \cite{WangICCV19}:
\begin{equation}
    \mcl = \sum_{ij} \mcl_{ij} = \sum_{ij} \texttt{BCE}(p_{ij}, x_{ij}).
\end{equation}

For pairwise matching pair $(\mcg_{a}, \mcg_{b})$, we can define the matching probability $p^{ab}_{ij}$ through universe graph. Let $p^{a}_{ik}$ denotes the universe matching probability between node $v_{i} \in \mcg_{a}$ and anchor point $v_{k}$ on universe graph $\mbfu$. Matching probability $p^{ab}_{ij}$ can be calculated as:
\begin{equation}
\begin{aligned}
    p^{ab}_{ij} &= \sum_{v_k \in \mbfu} p^{a}_{ik} p^{b}_{jk} = (p^{a}_{i})^{\top} p^{b}_{j},
    % p^{a}_{ik} &= \frac{\exp(\mcs^{a}_{ik})}{\sum_l \exp(\mcs^{a}_{il})}, \\
    % p^{b}_{jk} &= \frac{\exp(\mcs^{b}_{jk})}{\sum_l \exp(\mcs^{b}_{jl})}.
\end{aligned}
\end{equation}
where $p^{a}_{i}$ ($p^{b}_{j}$) is universe matching likelihood distribution between universe graph and $v_i$ ($v_j$).
They are obtained by normalization (\texttt{softmax}) on universe affinity $\mcs^{a}, \mcs^{b}$. 

We also calculate the gradient to analyze the effect of \texttt{BCE} loss on universe affinity:
\begin{equation}
    \frac{\partial \mcl_{ij}}{\partial \mcs^{a}_{it}} = 
    \left\{
        \begin{array}{cc}
            \frac{p_{it}^{a}}{p^{ab}_{ij}} \cdot \sum_{k} p_{ik}^{a} (p_{jk}^{b} - p_{jt}^{b}), & x_{ij} = 1; \\
            \frac{p_{it}^{a}}{p^{ab}_{ij}} \cdot \sum_{k} p_{ik}^{a} (p_{jt}^{b} - p_{jk}^{b}), & x_{ij} = 0. 
        \end{array}
    \right.
\end{equation} 
Moreover, the sum of the gradient on node $v_i$ would be 0:
\begin{equation}
    \sum_{t} \frac{\partial \mcl_{ij}}{\partial \mcs^{a}_{it}} = \pm \sum_{t} p_{it}^{a} \mp \frac{1}{p^{ab}_{ij}} \sum_{t} p_{it}^{a} p_{jt}^{b} = 1 - 1 = 0.
\end{equation}
The sign of gradient is decided by $\sum_{k} p_{ik}^{a} (p_{jk}^{b} - p_{jt}^{b})$. 
\begin{itemize} [leftmargin=*]
    \item When the ground truth $x_{ij} = 1$, $\mcs_{it}^{a}$ increases if $p_{jt}^{b}$ is large and $\mcs_{it}^{a}$ decreases if $p_{jt}^{b}$ is small. That is to say, the loss $\mcl_{ij}$ encourages the matching probability $p^{a}_{i}$ and $p^{b}_{j}$ getting closer.
    \item When the ground truth $x_{ij} = 0$, all the gradient are inverted, which means $p_{i}^{a}$ and $p_{j}^{b}$ are separated from each others.
\end{itemize}

\begin{algorithm2e}[tb!]
    \KwIn{Image pairs $\{\mci_1, \mci_2\}$, node coordinates $\{\mcv_1, \mcv_2\}$}
    
    Visual features are obtained from ResNet50 with relu3\_5 and relu4\_1 and final layers: $\mathcal{U}_{local}, \mathbf{F}_{middle}, \mathbf{G}_{global}  \gets \texttt{ResNet}(\mci_1, \mci_2, \mcv_1, \mcv_2)$ \;
    
    Node features are obtained as: $\mcf_{1}, \mcf_{2} = \texttt{MLP}(\mathcal{U}_{local}, \mathbf{F}_{middle},  \texttt{MLP}(\mathbf{G}_{global}))$ \;
    
    Node features are refined via Spline CNN with adjacency:
    $\mcf_{1}= \texttt{SplineCNN}(\mcf_{1}, \mca_{1})$ 
    $\mcf_{2} = \texttt{SplineCNN}(\mcf_{2}, \mca_{2})$ \;
    
    Calculate universe affinity for each graph:
    $\mcs_{1} \gets \texttt{SoftMax}(\texttt{MLP}(\mcf_{1}))$ 
    $\mcs_{2} \gets \texttt{SoftMax}(\texttt{MLP}(\mcf_{2}))$ \;
    
    \uIf{training phase}{
        Remove the last column to cope with outlier:
        $\mcs_{1}' \gets \mcs_{1}(:,:n-1)$, 
        $\mcs_{2}' \gets \mcs_{2}(:,:n-1)$ \;
        Calculate the matching loss: \hspace{2cm}
        $\mcl = \texttt{BCELoss}(\mcs_{1}' (\mcs_{2}')^{\top}, \mbfx_{gt})$
        
    } 
    \uElseIf{evaluation phase} {
        Remove the last column as well as outlier rows:
        $\mcs_{1}' \gets \texttt{OutlierFilter}(\mcs_{1}(:,:n-1))$
        $\mcs_{2}' \gets \texttt{OutlierFilter}(\mcs_{2}(:,:n-1))$ \;
        Calculate the matching to universe graph:
        $\mbfx_{1} \gets \texttt{Hungarian}(\mcs_{1}')$
        $\mbfx_{2} \gets \texttt{Hungarian}(\mcs_{2}')$ \;
        Fill the outlier rows with $0$ and reconstruct pairwise matching with $\mbfx_{1}, \mbfx_{2}$:
        $\mbfx_{12} \gets \texttt{Fill}(\mbfx_{1}) \texttt{Fill}(\mbfx_{2}^{\top})$
    }
    
    \KwOut{Pairwise matching $\mbfx_{12}$}
    \caption{Universe Partial Matching (\our)}
    \label{alg:main}
\end{algorithm2e}

\subsubsection{Outlier-aware loss design}
However, vanilla \texttt{BCE} loss can not handle outlier matching well. Since there is no correspondence between outliers $n_i$ and $n_j$, the universe matching likelihood distribution $p^{a}_i$ and $p^{b}_j$ are mutually exclusive. In that case, outliers can hardly be clustered and matched to the absorbing node in universe graph. To resolve it, we define the partial node matching probability by removing the last column (outlier column) of node matching probability $p_{ij}^{ab}$:
\begin{equation}
    p'_{ij} = \sum_{k=1}^{n-1} p^{a}_{ik} p^{b}_{jk},
\end{equation}
and the loss turns out to be:
\begin{equation}
    \mcl' = \sum_{ij} \texttt{BCE}(p'_{ij}, x_{ij}) = \sum_{ij} \texttt{BCE}(\sum_{k=1}^{n-1} p^{a}_{ik} p^{b}_{jk}, x_{ij})
\end{equation}

Under the new loss design, the gradient $\frac{\partial \mcl_{ij}'}{\partial \mcs^{a}_{it}}$ for $t \neq n$  does not change a lot:
\begin{equation}
\begin{small}
    \frac{\partial \mcl_{ij}'}{\partial \mcs^{a}_{it}} = 
    \left\{
        \begin{array}{cc}
            \frac{p_{it}^{a}}{p^{ab}_{ij}} \cdot [\sum_{k=1}^{n} p_{ik}^{a} (p_{jk}^{b} - p_{jt}^{b}) - p^{b}_{jn} p^{a}_{in}],&x_{ij} = 1 \\
            \frac{p_{it}^{a}}{p^{ab}_{ij}} \cdot [\sum_{k=1}^{n} p_{ik}^{a} (p_{jt}^{b} - p_{jk}^{b}) + p^{b}_{jn} p^{a}_{in}],&x_{ij} = 0 
        \end{array}
    \right.
\end{small}
\end{equation} 
whose sign is decided by $\sum_{k} p_{ik}^{a} (p_{jk}^{b} - p_{jt}^{b}) - p^{b}_{jn} p^{a}_{in}$, and all the conclusions above still hold. When $t = n$, the gradient becomes:
\begin{equation}
    \frac{\partial \mcl_{ij}'}{\partial \mcs^{a}_{in}} = 
    \left\{
        \begin{array}{cc}
            p_{in}^{a} p'_{ij} / p^{ab}_{ij}, & x_{ij} = 1 \\
            % \frac{p_{it}^{a}}{p_{ij}} \cdot (p_{ij} - p_{in}^{a} p_{jn}^{b}), & x_{ij} = 1 \\
            - p_{in}^{a} p'_{ij} / p^{ab}_{ij}, & x_{ij} = 0
            % -\frac{p_{it}^{a}}{p_{ij}} \cdot (p_{ij} - p_{in}^{a} p_{jn}^{b}), & x_{ij} = 0 
        \end{array}
    \right.
\end{equation}  

For outliers, there is no matching node and $x_{ij}$ is always 0. Thus the gradient is always negative and the affinity between outlier and absorbing node on universe graph will never decreases. 

For inliers, the universe matching likelihood distribution $p_{i}$ has two peaks: the correct anchor points and the absorbing node, where other terms are close to 0. Moreover, the gradient on $\mcs_{in}$ can be expanded into:
\begin{equation}
    - p_{in}^{a} p'_{ij} / p^{ab}_{ij} = - p_{in}^{a} \frac{\sum_{k=1}^{n-1} p_{ik}^{a} p_{jk}^{b}}{\sum_{k=1}^{n-1} p_{ik}^{a} p_{jk}^{b}}
\end{equation}
When the ground truth $x_{ij}=0$, only $p_{in}^{a} p_{jn}^{b}$ is relative large and other terms can be overlooked. Thus the gradient is close to 0.
On another side, when $x_{ij}=1$, both the matching likelihoods term of correct anchor points and the absorbing node are activated. Therefor $p'_{ij} / p^{ab}_{ij}$ is a constant instead of zero.
All in all, the gradient under positive match ($x_{ij}=1$) is much larger than the negative match ($x_{ij}=0$), which avoids inliers matching with absorbing node to some extent.

In general, by removing the last column (outlier column) in the node matching probability definition, the loss can distinguish the outlier and inlier. The outliers are also clustered into the absorbing node even without the guidance of the ground truth matching label, which allows for effective learning of inliers.

%%%%%%%%%%% Extension and Limitation %%%%%%%%%%%%%%
\subsection{Discussion on Limitation} \label{subsec:extension&limitation}

In face of the long-standing and challenging problem for partial matching of multiple graphs, our method \our \ also has several limitations. First, the definition of multiple graph correspondence (MGC) relies on the ground truth label of the dataset, which is used to distinguish between inlier and outlier. However, the credibility of this judgment depends on how much the graph pairs are labeled. When the dataset is small and noisy, or it is extremely sparsely labeled, the reliability of MGC is broken and thus leads to poor performance. As shown in `table' and `sofa' in Table~\ref{tab:voc_0outlier}, these two categories have few images and the performance becomes worse.

Another limitation is the relatively low recall of the prediction matching. Although our method \our \ can recognize the node type and thus successfully reduce the spurious matching, its prediction still leaves out many correct matching. As shown in Fig.~\ref{fig:matching_analysis}, the occurrence of ill-matching and over-matching decrease dramatically, while the appearance of correct matching does not increase a lot. 
That is to say, \our \ is cautious on its prediction and misses some match that is not obvious. It mainly attributes to the tendency that \our \ will mismatch some ill inliers to absorbing nodes. In some cases when a kind of inlier rarely appears over the whole dataset, the negative match's update surpasses the positive match's update and the inlier would be seen as an outlier. This problem prevents our method \our \ from making better predictions and leaves room for improvement in future work.

%%%%%%%%%%%%%%%%%%%%%%%%%%%%%%%%%% Experiments %%%%%%%%%%%%%%%%%%%%%%%%%%%%%%%%%%%%%%%%%%%%%%%%%%%%%%%%%%%%%%%%
\section{Experiments}

In this section, we first introduce the general protocol, metric, and dataset processing for all experiments in Sec.~\ref{subsec:protocol}. Then we introduce peer methods in Sec.~\ref{subsec:peer_method}. In Sec.~\ref{subsec:PM}, \ref{subsec:exploration} and \ref{subsec:extension_case}, we show experiments we have done from three perspectives: partial matching comparison, method exploration \& analysis, and extension cases exploration. All the experiments are conducted on a Linux workstation with Nvidia 2080Ti GPU and Intel i7-7820X CPU @ 3.60GHz with 128GB RAM.

\subsection{Protocol, Metrics, and Dataset} \label{subsec:protocol}

\vspace{5pt}
\subsubsection{Training and Evaluation Protocol}
We adopt three protocols for our experiments: standard partial matching, online matching, and mixture graph matching and clustering.

\begin{itemize} [leftmargin=*]
    \item \textbf{Standard partial matching} is the fundamental setting in this paper. The unmatched inliers and outliers are either obtained from the dataset itself or manually added. In some experiments, there might be no unmatched inlier or outlier. The training graphs are randomly sampled in the same class and we only sample two graphs once a time to build the matching instance. The evaluation phase also follows the training protocol.
    \item \textbf{Online matching} is explored by \cite{YuECCV18, JiangPAMI21} where input graphs are not given at once, but come one by one. In this paper, we adopt the protocol of standard partial matching for the training phase. While in the evaluation phase, we sample 15 graphs of the same category and send them into the model one after another. Moreover, the online matching is conducted on partial matching, where unmatched inliers and outliers are also added following standard partial matching's protocol.
    \item \textbf{Mixture graph matching and clustering} is proposed by \cite{WangAAAI20}. It aims to deal with the input of different classes and obtain both matching and clustering results. We also add outliers for the dataset as the first two settings do. In the training phase, we keep half of the samples in the same class and other samples consist of graphs from different classes. The ground truth matching is set to zero between inter-class graphs. In the evaluation phase, we follow the protocol in \cite{WangAAAI20, wang2020graduated} to randomly pick several classes and sample graphs in selected classes. The clustering results are obtained via spectral clustering~\cite{ng2002spectral} based on affinity score. The affinity score of \our \ is defined as:
    \begin{equation}
        \text{Affinity} = \hat \mbfx_{ab} \circ (\mcs_{a} \mcs_{b}^{\top})
    \end{equation}
    where $\circ$ denotes the Hadamard product. 
\end{itemize}

\subsubsection{Evaluation Metric} 
We use multiple metrics for performance evaluation as it is beyond the full  two-graph matching setting.
%\hspace{3pt}

\textbf{F1 Score} \ We seem the matching prediction as a binary classification task for each entry. Therefore, F1 score is applied as an evaluation metric, which is first used by BBGM~\cite{rolinek2020deep} in `Keypoint inclusion' setting. The F1 score is calculated as:
\begin{equation}
    \mathbf{F}_1 = 2 \cdot \frac{\text{Precision} \cdot \text{Recall}}{\text{Precision} + \text{Recall}} = \frac{\text{TP}}{\text{TP} + \frac{1}{2}(\text{FP}+\text{FN})}
\end{equation}
where TP denotes true positive, FP denotes false positive and FN denotes false negative. F1 score is the most important metric in our experiments.

\textbf{Accuracy.} \ We also adopt the accuracy as evaluation metric in online matching, which represents the correct node pairs' ratio in prediction matching:
\begin{equation}
    \mathbf{acc} = 1 - \frac{\|\bar \mbfx - \mbfx_{gt}\|^2_F}{\| \bar \mbfx \|^2_F}
\end{equation}
where $\| \cdot \|_F$ denotes Frobenius Norm. 

\begin{table}[tb]
    \caption{\textbf{Wrong match types:} mismatching, ill-matching, and over-matching. The wrong match types are defined by the node type of two nodes that belong to the specific node correspondence. The header row and header column represent their node types. Moreover, \textbf{MI}, \textbf{UI}, and \textbf{O} refer to matched inlier, unmatched inlier, and outlier respectively.  }
    \label{tab:match_type}
    %\vspace{-5pt}
    \resizebox{0.49\textwidth}{!}{
    \renewcommand\arraystretch{1.2}
    \begin{tabular}{|c|c|c|c|}
        \hline
                         & \textbf{MI} & \textbf{UI} & \textbf{O}  \\ \hline
        \textbf{MI}   & mismatching       & ill-matching        & over-matching  \\ \hline
        \textbf{UI} & ill-matching      & ill-matching        & over-matching \\ \hline
        \textbf{O}         & over-matching     & over-matching       & over-matching \\ \hline
    \end{tabular}}
   % %\vspace{-20pt}
\end{table}

\textbf{Matching Type.} \ Matching type is defined to illustrate the ability of node type recognition.  We divide all the matching predictions into four types: correct matching, mismatching, ill-matching, and over-matching. The first one denotes the correct prediction whereas the last three denotes three different types of errors. The matching type of prediction $(v_i, v_j)$ is defined by the node type of $v_i$ and $v_j$, as shown in Table~\ref{tab:match_type}. The number of ill-matching denotes the ability of the model to distinguish unmatched inliers, and the number of over-matching denotes the ability of the model to distinguish outliers. We calculate the ratio of each matching type to show the distinguishing ability of node types.

\textbf{Clustering Metrics.} \ Clustering Purity, Rand Index,  Clustering Accuracy and Matching Accuracy~\cite{WangAAAI20} are also introduced to evaluate the clustering results of mixture graph matching and clustering in Sec.~\ref{subsec:extension_case}:
\begin{itemize}[leftmargin=*]
    \item Clustering Purity (\textbf{CP}): Let $\mcc_i$ represents the $i$-th predicted cluster, $\mcc_j^{gt}$ is the $j$-th ground truth cluster, $k$ denotes the number of cluster, and $m$ denotes the number of total graphs.
    \begin{equation}
        \textbf{CP} = \frac{1}{m} \sum_{i=1}^{k} \max_{j \in \{1 \dots k\}} | \mcc_i \cap \mcc_j^{gt}|.
    \end{equation}
    \item Rand Index (\textbf{RI}): it is computed by the number of graphs predicted in the same cluster with the same label $n_{11}$, and the number of graphs predicted in separate clusters and with different labels $n_{00}$, and normalized by the total number of graph pairs $n$.
    \begin{equation}
        \textbf{RI} = \frac{n_{11} + n_{00}}{n}.
    \end{equation}
    \item Clustering Accuracy (\textbf{CA}): the accuracy of cluster prediction, is measured by calculating all the wrong match pairs from same or different clusters.
    \begin{equation}
    \begin{aligned}
        \textbf{CA} = 1 - \frac{1}{k} (&\sum_{\mcc_i} \sum_{\mcc_{j_1}^{gt} \neq \mcc_{j_2}^{gt}} \frac{1}{|\mcc_i||\mcc_i|} \cdot |\mcc_i \cap \mcc_{j_1}^{gt}| |\mcc_i \cap \mcc_{j_2}^{gt}| \\ +& \sum_{\mcc_{i_1} \neq \mcc_{i_2}} \sum_{\mcc_{j}^{gt}} \frac{1}{|\mcc_{i_1}||\mcc_{i_2}|} \cdot |\mcc_{i_1} \cap \mcc_{j}^{gt}| |\mcc_{i_2} \cap \mcc_{j}^{gt}|)
    \end{aligned}
    \end{equation}
    \item Matching Accuracy of Cluster / Matching F1 Score of Cluster (\textbf{MAC/F1C}): the matching quality within each cluster.
    \begin{equation} \label{eq:ma}
        \begin{aligned}
        \textbf{MA} &= 1 - \frac{1}{k} \sum_{j=1}^{k} \frac{1}{|\mcc_{j}|^2} \sum_{(\mcg_{a}, \mcg_{b}) \in \mcc_{j}} \textbf{acc}(\mbfx_{ab}, \mbfx^{gt}_{ab}) \\
        \textbf{F1C} &= 1 - \frac{1}{k} \sum_{j=1}^{k} \frac{1}{|\mcc_{j}|^2} \sum_{(\mcg_{a}, \mcg_{b}) \in \mcc_{j}} \textbf{F1}(\mbfx_{ab}, \mbfx^{gt}_{ab}) 
        \end{aligned}
    \end{equation}
\end{itemize}

\begin{table*}[tb!]
    \caption{F1 score comparison on PascalVOC with all unmatched inliers and two random outliers. All the learning-based methods apply the ResNet backbone. The affinity matrix of the learning-free solver is built with the unary and quadratic cost learned by BBGM. We either adopt the optimal hyperparameters reported in their paper or adjust them to fit the partial matching setting. }
    \label{tab:voc_2outlier}
    %\vspace{-5pt}
    \resizebox{\textwidth}{!}{
    \renewcommand\arraystretch{1.4}
    \begin{tabular}{|c|cccccccccccccccccccc|c|}
        \hline
        \textbf{Methods} & \textbf{aero} & \textbf{bike} & \textbf{bird} & \textbf{boat} & \textbf{bottle} & \textbf{bus}  & \textbf{car}  & \textbf{cat}  & \textbf{chair} & \textbf{cow}  & \textbf{table} & \textbf{dog}  & \textbf{horse} & \textbf{motor} & \textbf{person} & \textbf{plant} & \textbf{sheep} & \textbf{sofa} & \textbf{train} & \textbf{tv}   & \textbf{avg}   \\ \hline
        \textbf{IPCA~\cite{WangICCV19}}             & 32.6          & 51.0          & 32.7          & 26.4          & 64.0          & 45.2          & 25.1          & 43.0          & 26.0          & 41.3          & 22.1          & 39.5          & 41.0          & 40.1          & 30.7          & 57.6          & 33.0          & 23.1          & 37.0          & 58.1          & 38.46          \\
        \textbf{NGM~\cite{Wang2019Neural}}           & 35.3          & 57.4          & 42.3          & 31.9          & 68.5          & 47.0          & 30.7          & 49.4          & 33.4          & 48.3          & 38.5          & 48.5          & 46.7          & 49.5          & 39.9          & 70.8          & 36.8          & \textbf{38.7}          & 45.0          & 58.1          & 45.84          \\
        \textbf{BBGM~\cite{rolinek2020deep}}             & 36.2          & 63.2          & 54.1          & 36.0          & 75.1          & 60.8          & 20.9          & 65.1          & 33.7          & 61.2          & 36.0          & 60.1          & 57.4          & 57.2          & 39.4          & 81.3          & 53.5          & 27.1          & 51.3          & 78.5          & 52.41          \\ \hline
        \textbf{GAGM~\cite{Gold1996AGA}}             & 26.7          & 49.3          & 36.6          & 25.1          & 61.9          & 40.7          & 9.0           & 45.4          & 22.5          & 39.3          & 13.1          & 41.0          & 37.8          & 39.1          & 27.9          & 59.8          & 29.1          & 19.7          & 26.0          & 54.9          & 35.25          \\
        \textbf{RRWM~\cite{Cho2010ReweightedRW}}             & 28.3          & 54.6          & 40.3          & 27.4          & 70.6          & 46.3          & 9.0           & 49.8          & 23.5          & 42.8          & 15.1          & 44.8          & 38.1          & 39.1          & 30.3          & 66.8          & 29.3          & 18.1          & 27.1          & 64.0          & 38.27          \\
        \textbf{BPF~\cite{Wang2018GraphMW}}              & 25.8          & 55.2          & 38.8          & 26.0          & 70.7          & 45.7          & 9.5           & 49.4          & 22.9          & 40.2          & 14.4          & 44.1          & 36.9          & 40.0          & 29.0          & 68.6          & 28.6          & 18.8          & 28.1          & 63.8          & 37.83          \\
        \textbf{ZAC~\cite{wang2020zero}}              & 37.2          & 62.3          & 50.8          & 34.0          & 76.0          & 55.9          & 23.8          & 63.7          & 33.2          & 58.8          & 37.7          & 59.4          & 55.8          & 55.4          & 39.1          & 76.8          & 49.3          & 29.0          & 49.1          & 68.9          & 50.81          \\ \hline
        \our & \textbf{46.3} & \textbf{67.2} & \textbf{65.3} & \textbf{44.9} & \textbf{80.5} & \textbf{72.7} & \textbf{63.5} & \textbf{71.9} & \textbf{49.1} & \textbf{73.5} & \textbf{58.8} & \textbf{69.4} & \textbf{63.6} & \textbf{62.4} & \textbf{55.5} & \textbf{86.4} & \textbf{63.0} & 28.4 & \textbf{68.5} & \textbf{80.8} & \textbf{63.57} \\ \hline
    \end{tabular}}
   % %\vspace{-10pt}
\end{table*}

\subsubsection{Dataset Processing} \label{subsec:dataset_and_metric}
\vspace{5pt}
\noindent \textbf{PascalVOC.} \hspace{3pt}
The Pascal VOC~\cite{PascalVOC} with Berkeley annotations ~\cite{Bourdev2009PoseletsBP} contains images with bounding boxes surrounding objects of 20 classes. We follow the standard data preparation procedure of NGM and BBGM~\cite{Wang2019Neural, rolinek2020deep}. Each object is cropped to its bounding box and scaled to $256 \times 256$ px. The resulting images contain up to 23 annotated key points, depending on the object category. We follow the `Keypoint inclusion' setting proposed in \cite{rolinek2020deep} which does not filter out unmatched inliers for graph pairs. Moreover, we randomly sample the coordinates on images to add outliers for each class. We also follow NGM~\cite{Wang2019Neural} and BBGM~\cite{rolinek2020deep} to split the train and test dataset, where training data includes 7,020 images and testdata includes 1,682 images. 

\vspace{5pt}
\noindent \textbf{Willow ObjectClass.} \hspace{3pt}
The Willow ObjectClass~\cite{Cho2013LearningGT} contains images from Caltech-256~\cite{caltech256} and Pascal VOC 2007~\cite{everingham2007pascal}, which consists of 256 images from 5 categories: 40 cars, 40 motorbikes, 50 ducks, 66 wine bottles, and 109 faces.  Each image is annotated with the same 10 distinctive category-specific key points. Following standard procedure, we also crop the images to the bounding boxes of the objects and rescale them to $256 \times 256$ px. In experiments, we randomly drop some key points for each image to generate unmatched inliers. We also add random outliers as we do in PascalVOC. We choose 20 images from each category as our training dataset and leave others for evaluation.

\subsection{Compared Methods} \label{subsec:peer_method}
We compare our methods \our \ with previous work of two kinds: deep graph matching learning model and traditional solver with pre-defined affinity. In our experiments, the affinity in the second category is pre-learned to make the comparison as fair as possible.

\subsubsection{Learning-based model}

Unlike previous works mostly adopting VGG~\cite{simonyanICLR14vgg} as the backbone, in our experiment, we apply ResNet50~\cite{he2016deep} to improve the model's expression capability and cost-efficiency. For a fair comparison, we replace the backbone with ResNet50 for all previous works mentioned below and train them with optimal hyperparameters released in their codes. The feature dimension is reduced to 1024 with MLP layers as the same with \our. 
\begin{itemize}[leftmargin=*]
    \item \textbf{PCA / IPCA} \cite{WangICCV19} proposes a method called permutation loss and cross-graph affinity graph matching model (PCA-GM) which is the first deep learning model that applies GNN and permutation loss. \cite{WangPAMI20} further improved the idea with the method called Iterative Permutation loss and Cross-graph Affinity based Graph Matching (IPCA-GM).
    \item \textbf{NGM} \cite{Wang2019Neural} proposes Neural Graph Matching network (NGM) to address the most general Lawler’s QAP form via a learning-based solver. NGM construct an association graph based on the affinity matrix and transfer the assignment problem into the node classification task on the association graph.
    \item \textbf{BBGM} \cite{rolinek2020deep} proposes BlackBox differentiation of Graph Matching solvers (BBGM) to combine the Spline CNN refinement module and traditional graph matching algorithm with blaxkbox differentiation.
    \item \textbf{DLGM} \cite{yu2021deep} focuses on exploring graph topology with Deterministic model (DLGM-D) and generative model (DLGM-G) to improve matching quality. We report the performance of DLGM-G as it outperforms DLGM-D in all kinds of experiments in the paper ~\cite{yu2021deep}. Note we do not replace its backbone with ResNet50 and we report its number which is obtained from \cite{yu2021deep} for it has not released code yet.
\end{itemize}

\subsubsection{Learning-free solver}

\begin{table*}[tb!]
    \caption{F1 score comparison on PascalVOC with all unmatched inliers and without any outlier. All the learning-based methods apply the ResNet backbone except DLGM.  Note DLGM has not released its code. So we report its number reported by the paper, which still uses the VGG16 backbone. The affinity matrix of the learning-free solver is built with the unary and quadratic cost learned by BBGM. We either adopt the optimal hyperparameters reported in their paper or adjust them to fit the partial matching setting.}
    \label{tab:voc_0outlier}
    %\vspace{-5pt}
    \resizebox{\textwidth}{!}{
    \renewcommand\arraystretch{1.4}
    \begin{tabular}{|c|cccccccccccccccccccc|c|}
        \hline
         \textbf{Methods} & \textbf{aero} & \textbf{bike} & \textbf{bird} & \textbf{boat} & \textbf{bottle} & \textbf{bus}  & \textbf{car}  & \textbf{cat}  & \textbf{chair} & \textbf{cow}  & \textbf{table} & \textbf{dog}  & \textbf{horse} & \textbf{motor} & \textbf{person} & \textbf{plant} & \textbf{sheep} & \textbf{sofa} & \textbf{train} & \textbf{tv}   & \textbf{avg}   \\ \hline
        \textbf{IPCA~\cite{WangICCV19}}    & 33.0          & 56.5          & 38.8          & 32.1          & 80.1            & 55.0          & 30.8          & 43.4          & 32.2           & 44.8          & 39.1           & 40.5          & 44.2           & 52.4           & 34.0            & 81.0           & 38.8           & 26.1          & 55.7           & 75.3          & 46.69          \\
        \textbf{NGM~\cite{Wang2019Neural}}   & 42.4          & 69.3          & 51.2          & 40.5          & 87.0            & 62.6          & 43.5          & 57.9          & 45.1           & 57.3          & 53.4           & 55.0          & 55.9           & 62.5           & 46.6            & 92.5           & 44.7           & 41.8          & 66.6           & 73.6          & 57.46          \\
        \textbf{DLGM*~\cite{yu2021deep}}    & 43.8          & 72.9          & 58.5          & \textbf{47.4} & 86.4            & 71.2          & 53.1          & 66.9          & 54.6           & 67.8          & \textbf{64.9}           & 65.7          & 66.9           & 70.8           & 47.4            & 96.5           & 61.4           & \textbf{48.4} & 77.5  & 83.9          & 65.30          \\
        \textbf{BBGM~\cite{rolinek2020deep}}    & 45.7          & 75.4          & 64.3          & \textbf{47.4} & \textbf{88.2}   & 63.3          & 49.5          & 70.2          & 44.9           & 69.5          & 54.8  & 67.3          & 66.6           & 71.8           & 53.1            & 95.9           & 67.0           & 33.0          & \textbf{80.5}           & 80.5          & 64.45          \\  \hline
        \textbf{GAGM~\cite{Gold1996AGA}}    & 37.3          & 60.3          & 47.7          & 39.1          & 75.0            & 53.7          & 36.1          & 59.2          & 34.8           & 56.2          & 40.2           & 54.4          & 51.9           & 54.3           & 39.0            & 83.8           & 48.8           & 19.0          & 58.3           & 68.7          & 50.89          \\
        \textbf{RRWM~\cite{Cho2010ReweightedRW}}    & 39.4          & 63.7          & 52.1          & 39.7          & 76.5            & 57.9          & 28.5          & 64.3          & 37.9           & 60.3          & 43.0           & 57.9          & 54.2           & 54.0           & 43.3            & 85.0           & 50.4           & 22.0          & 66.1           & 69.9          & 53.30          \\
        \textbf{BPF~\cite{Wang2018GraphMW}}     & 39.2          & 66.5          & 51.2          & 40.0          & 77.0            & 59.0          & 26.7          & 63.8          & 36.9           & 60.6          & 49.7           & 58.1          & 54.3           & 58.2           & 43.6            & 84.9           & 50.7           & 23.0          & 66.4           & 70.4          & 54.02          \\
        \textbf{ZAC~\cite{wang2020zero}}     & 43.5          & 62.8          & 57.5          & 43.9          & 72.1            & 61.2          & 33.6          & 69.0          & 38.3           & 64.1          & 46.4           & 66.2          & 61.8           & 61.4           & 46.4            & 83.2           & 58.1           & 27.9          & 69.8           & 72.6          & 56.99          \\  \hline
        \our & \textbf{50.5} & \textbf{74.1} & \textbf{67.3} & 46.1          & 86.5            & \textbf{74.9} & \textbf{75.6} & \textbf{73.0} & \textbf{59.4}  & \textbf{73.0} & 43.3           & \textbf{72.4} & \textbf{67.6}  & \textbf{72.2}  & \textbf{55.2}   & \textbf{97.5}  & \textbf{68.3}  & 45.0          & 75.5           & \textbf{85.1} & \textbf{68.11} \\  \hline
    \end{tabular}}
   % %\vspace{-15pt}
\end{table*}

learning-free solvers only take the affinity and do not have models like front-end feature extraction and matching instance building to extract information from the real world. Therefore, we adopt the learned affinity extracted by BBGM~\cite{rolinek2020deep} with ResNet50 backbone as input for all learning-free solvers. All the hyperparameters are set to the optimal as they reported.
\begin{itemize}[leftmargin=*]
    \item \textbf{GAGM} \cite{Gold1996AGA}  proposes a graduated assignment algorithm for graph matching which is fast and accurate even in the presence of high noise. This classic method has inspired a series of GM solvers including \cite{YanCVPR15,TianECCV12}.
    \item \textbf{RRWM} \cite{Cho2010ReweightedRW} adopts a random-walk view with reweighted jump on association graph based on affinity matrix.
    \item \textbf{BPF} \cite{Wang2018GraphMW} improves path following techniques by branch switching, reaching state-of-the-art performance on learning-free solvers of graph matching.
    \item \textbf{ZAC} \cite{wang2020zero} works on KAP formulation and designs an efficient outlier-robust algorithm to significantly reduce the incorrect or spurious matchings caused by numerous outliers.
    \item \textbf{CAO} \cite{YanPAMI16} propose Composition Based Affinity Optimization (CAO-c) to solve MGM via iteratively update pairwise matching with new matching composition. It also proposed two methods CAO-pc and CAO-uc to accelerate the original method.
    \item \textbf{DPMC} \cite{WangAAAI20} first propose Decayed Pairwise Matching Composition based method, called DPMC, to solve the mixture graph matching and clustering problem. It adopts a tree-structured supergraph with affinity score to distinguish different classes.
    \item \textbf{GA-MAMC / GANN-MGMC} \cite{wang2020graduated} propose GA-MAMC, an MGM solver with graduate assignment, where GANN-MGMC is its self-supervised learning version.
\end{itemize}

\begin{figure*}[tb!]
	\centering
    \includegraphics[width=\textwidth]{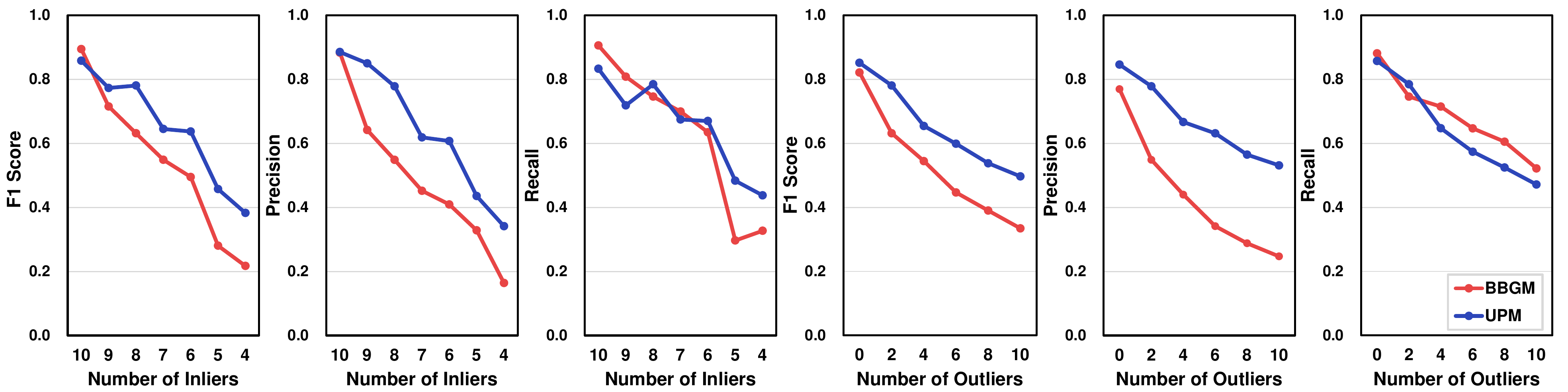}
 	\caption{F1 score, precision and recall on Willow-Object with different extensions of partial matching cases. The figures on the left show the robustness of our method against unmatched inliers, while the figures on the right show the robustness of our method against outliers.}
 	\label{fig:willow_occlusion&outlier}
\end{figure*}

\begin{figure}[tb!]
    \centering
    \includegraphics[width=0.49\textwidth]{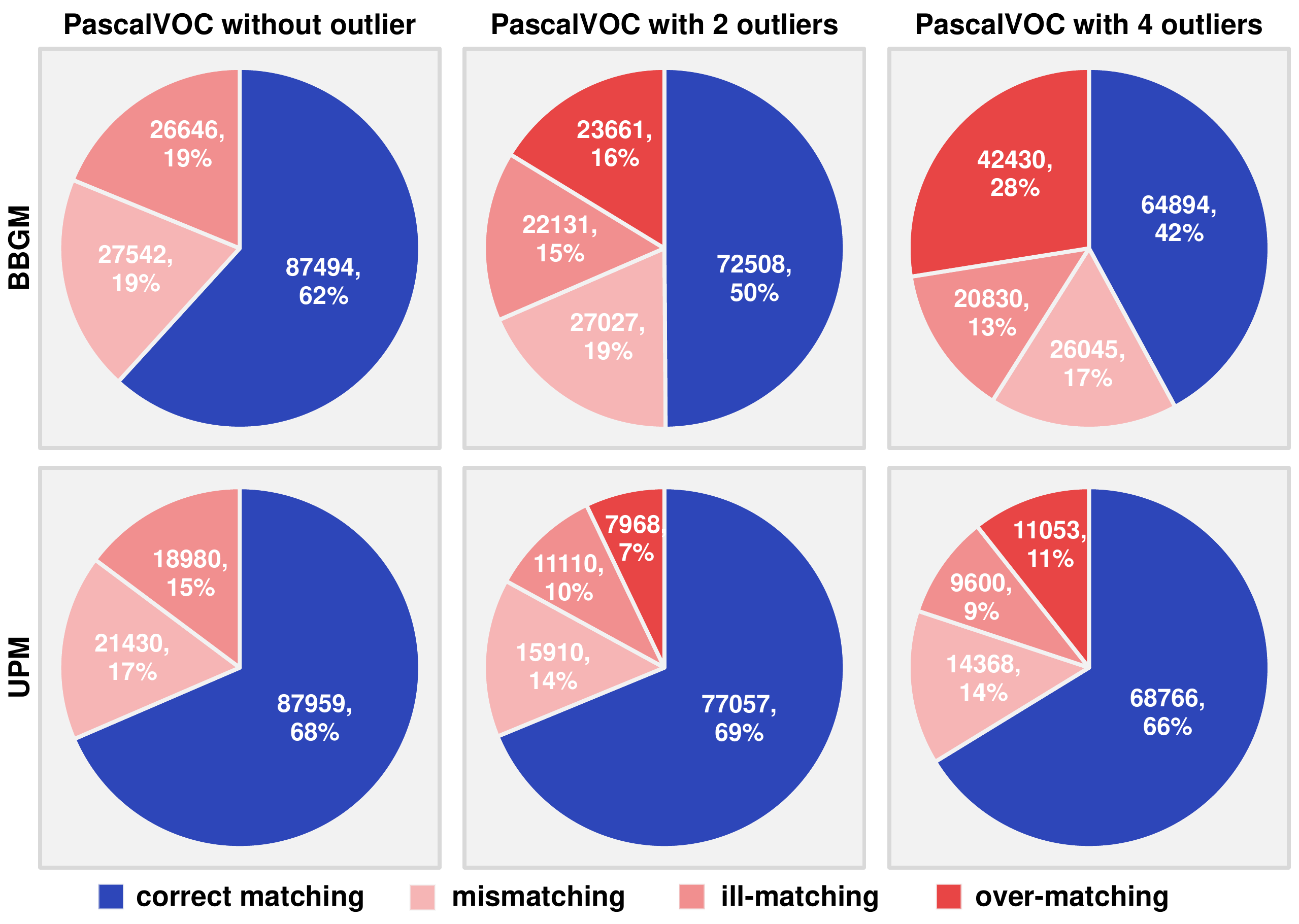}
   % %\vspace{-10pt}
    \caption{Distribution of matching type on PascalVOC with different numbers of outliers. For representativeness, both state-of-the-art BBGM and the proposed \our \ are tested. The occurrence number of each matching type over all the categories are summed up. Both the occurrence number and proportion are reported for each matching type. It shows our method \our \ has superiority on distinguishing node types.}
    \label{fig:matching_analysis}
   % %\vspace{-5pt}
\end{figure}

\subsection{Partial Matching Performance} \label{subsec:PM}
We conduct two experiments for standard partial matching comparison on PascalVOC with different kinds of illness: 
\begin{itemize}[leftmargin=*]
    \item PascalVOC with unmatched inliers and 2 outliers,
    \item PascalVOC with unmatched inliers and without outlier.
\end{itemize}
During the training phase, we train \our \ for 10 epochs, and in each epoch, we random sample $16000$ pairs of graphs for training. We also randomly sample $8000$ pairs for evaluation. We apply the feature-merged universe graph of size 30 and feature dimension 1024. The F1 score is reported in Table~\ref{tab:voc_0outlier} and Table~\ref{tab:voc_2outlier}.

As one can see, our method \our \ outperforms all the peer methods in both settings. The improvement is large in the complex settings where \our \ is 11.16\% and 12.76\% higher than the sota of both deep learning model BBGM and learning-free solver ZAC on average, and even outperforms the peer methods across almost all the categories. 
As for the setting without outlier, modules like outlier-aware loss and outlier filter do not work anymore. However, the gap between \our \ and other methods is  still remarkable: 2.81\% (for learning methods) and 11.12\% (for learning free solvers). 
Moreover, our method \our \ has really bad performance on `table' (in Table~~\ref{tab:voc_0outlier}) and `sofa' (in Table~~\ref{tab:voc_2outlier}), which supports our analysis in Sec.~\ref{subsec:extension&limitation} that our method can suffer from small and noisy dataset.

% The distribution of matching type is shown in Fig.~\ref{fig:matching_analysis} for the setting of different outliers. For BBGM, the ratio of over-matching increases dramatically with adding the outliers, while it is almost unchanged in our method \our. The ratios of all the kinds of wrong matching types are lower than BBGM, which shows that our model \our \ has advantages on distinguishing node type.

Fig.~\ref{fig:matching_analysis} shows the comparison on node type recognition. Three experiments with 0/2/4 outliers are conducted and the matching type ratio (proposed in Tab.~\ref{tab:match_type}) of both BBGM and \our \ are reported. As one can see, \our \ achieves 6\% improvements on correct matching ratio under the setting without outlier. The gap is increasing by adding the outliers, which reaches 24\% under 4 outliers. Meanwhile, the wrong match ratio of \our \ keeps the same when the outlier number increases from 0 to 4. It illustrates that \our \ has a robust performance against outlier, and we will show that with more rigorous experiments in the following.

We also conduct two pressure tests of learning methods on Willow Object for unmatched inlier and outlier respectively:
\begin{itemize}[leftmargin=*]
    \item Willow Object with 2 outliers, random drop $0 \sim 6$ inliers,
    \item Willow Object with $0 \sim 10$ outliers, random drop $2$ inliers.
\end{itemize}
We train the model for 2 epochs, where 400 and 100 pairs are randomly sampled for training and evaluation respectively. We also set the universe graph size as 20 or 25 and feature dim as 1024 to keep feature-merged universe graph construction for \our.

\begin{figure*}
    \centering
    \includegraphics[width=0.85\textwidth]{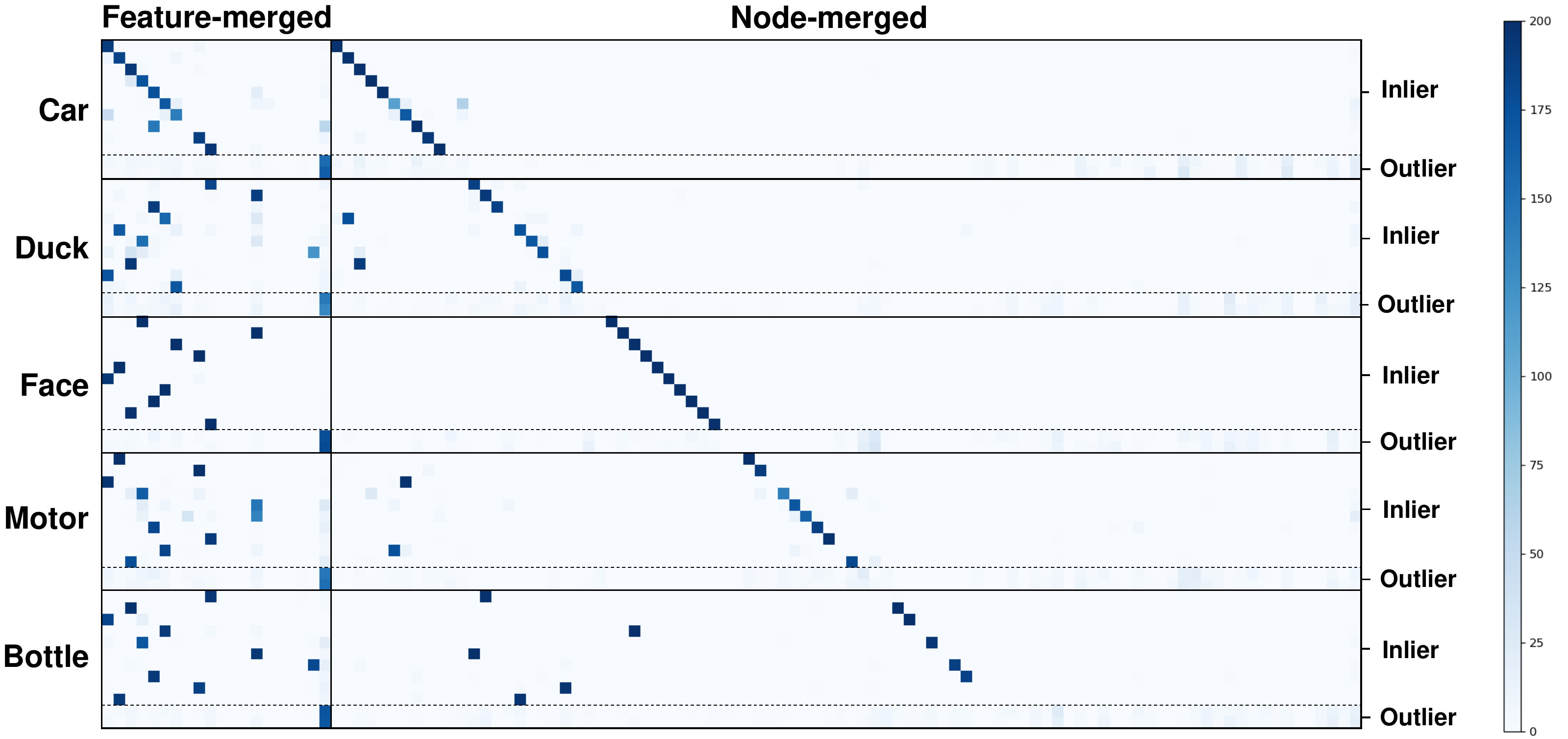}
    \caption{Universe node matching likelihood distribution on Willow Object with two outliers. We randomly sample 200 graphs for each category and count the correspondence between input graphs and two types of universe graphs. The X-axis denotes the nodes in the feature-merged graph and node-merged universe graph, whose size is $20$ and $90$ respectively. The Y-axis denotes the nodes of different semantic meaning in all the categories, where inlier and outlier are divided by dotted lines. Darker color represents more appearances for the node correspondence pair. Since all the universe nodes are permutation invariant, we adjust the order of columns for better visualization.}
    \label{fig:node_distribution}
\end{figure*}

\begin{figure*}[tb!]
    \centering
    \includegraphics[width=\textwidth]{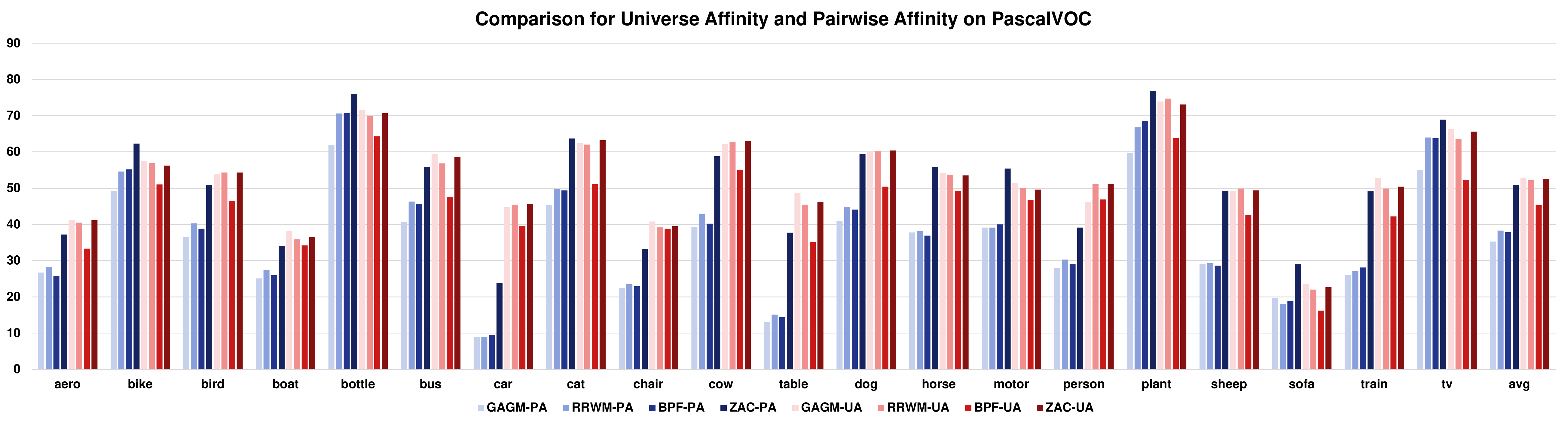}
   % %\vspace{-10pt}
    \caption{Comparison of F1 score for universe affinity and pairwise affinity on PascalVOC. The average F1 score is reported at last. Universe affinity matrix $\mbfk^{u}$ is built with $\mcs$ learned by \our\ as shown in Eq.~\ref{eq:universe_affinity_matrix}. Pairwise affinity matrix is learned by BBGM with Resnet50 backbone. Methods with `-UA' denotes the learning-free solver with universe affinity $\mbfk^{u}$, and `-PA' denotes for pairwise affinity matrix.}
    \label{fig:universe_affinity}
\end{figure*}

As shown in Fig.~\ref{fig:willow_occlusion&outlier}, our method \our \ outperforms BBGM, especially in extreme cases. In the inlier test, \our \ outperforms BBGM with 5\%-10\% improvements on F1 score when inlier number is less than 10 and there exists unmatched inlier. In the outlier test, \our \ beats BBGM on F1 score with 3\% - 15\% in all the settings. More complex the setting is, \our \ is more powerful to achieve better matching results. Meanwhile, it also shows that \our \ has a great advantage on the precision of the prediction, where the recall of \our \ is almost the same or lower than BBGM in all the cases. This also supports our discussion in Sec.~\ref{subsec:extension&limitation} that \our \ tends to give low confidence matching prediction.

\subsection{Further Exploration on \our} \label{subsec:exploration}
In this subsection, we conduct four experiments on: universe graph structure, quality of universe affinity, generalization ability, and hyper-parameter sensitivity to further analyze our algorithm \our.

\subsubsection{Universe graph structure}
To confirm our hypothesis with universe graph structure we've talked in Sec.~\ref{subsec:pipeline}, we conduct an experiment on Willow Object and draw the node map distribution of feature-merged universe graph and node-merged universe graph, as shown in Fig.~\ref{fig:node_distribution}. 

As we mentioned before, the feature-merged universe graph is constructed with small size \our-\textbf{20} (20 denotes the universe graph size $n_u$), and the node-merged universe graph is built with \our-\textbf{90} on the opposite. We train \our-\textbf{20} with standard partial matching setting and train \our-\textbf{90} with mixture graph matching setting. Both models are trained with two outliers and no inlier is dropped. \our-\textbf{20} and \our-\textbf{90} all achieve $85$ on F1 score under their setting, which means the model is almost ideally trained according to our theory. We show the node map distribution to show the structure of two universe graphs in Fig.~\ref{fig:node_distribution}. 

First of all, the feature-merged graph concatenates nodes feature from different classes for different nodes in different classes are matched to the same universe node. Meanwhile, the node-merged universe graph divides all the nodes separately. This supports our hypothesis on the structure of these two graphs, which is proposed in Fig.~\ref{fig:universe_graph}. Furthermore, the outlier distribution in the feature-merged universe graph also confirms our design on outlier filter and loss modules. The outlier is matched to the last node, called absorbing node, on the universe graph. As for the node-merged universe graph, the outliers are scattered from 61-th anchor node to 90-th anchor node, which is mainly attributed to the universe size exceeding the sum of exact feature number a lot.

\subsubsection{Universe affinity vs pairwise affinity} \hspace{3pt}
As we mentioned in Eq.~\ref{eq:ua2pa}, universe affinity does not lose information compared to pairwise affinity, since the latter can be built with the former. Therefore, we conduct a experiment to show the quality of pairwise affinity matrix constructed by universe affinity. Since we only have node affinity, the affinity matrix $\mbfk^{u}$ between $\mcg_a$ and $\mcg_b$ is calculated as:
\begin{equation} \label{eq:universe_affinity_matrix}
    \begin{aligned}
        \mbfk^{u}_{n_{i}, n_{k}} &= \mcs_{ab}(i,k)^2 \\
        \mbfk^{u}_{e_{ij}, e_{kl}} &= \mcs_{ab}(i,k) \cdot \mcs_{ab}(j,l)
    \end{aligned}
\end{equation}
where $\mcs_{ab}$ is pairwise affinity built by universe affinity $\mcs_a, \mcs_b$ in Eq.~\ref{eq:ua2pa}. It is well-defined since in bijective matching setting with fully connected graph topology, the quadratic assignment problem on $\mbfk^{u}$ equals to linear assignment problem on $\mcs_{ab}$.
\begin{equation}
    \begin{aligned}
        & \argmax_{\mbfx} \vct(\mbfx)^\top \mbfk^{u} \vct(\mbfx) \\
        % = & \argmax_{\mbfx} \sum_{\mbfx_{ik} = 1} \sum_{\mbfx_{jl} = 1} \mbfx_{e_{ij}, e_{kl}} \\
        = & \argmax_{\mbfx} \sum_{\mbfx_{ik} = 1} \sum_{\mbfx_{jl} = 1} \mcs_{ab}(i,k) \cdot \mcs_{ab}(j,l) \\
        % = & \argmax_{\mbfx} \bigg ( \sum_{\mbfx_{ik} = 1} \mcs_{ab}(i,k) \bigg ) \cdot \bigg ( \sum_{\mbfx_{jl} = 1}  \cdot \mcs_{ab}(j,l) \bigg ) \\
        = & \argmax_{\mbfx} \bigg ( \sum_{\mbfx_{ij} = 1} \mcs_{ab}(i,j) \bigg )^2 \\
        =& \argmax_{\mbfx} \sum_{\mbfx_{ij} = 1} \mcs_{ab}(i,j) \\
    \end{aligned}
\end{equation}

We compare universe affinity $\mbfk^{u}$ with the pairwise affinity $\mbfk$ learned by BBGM with ResNet50 backbone. They are fed into several learning-free solvers: GAGM, RRWM, BPF, and ZAC for comprehensive evaluation. We follow the standard partial matching setting on PascalVOC without filtering out any unmatched inlier and randomly add two outliers for each image. The F1 score is reported in Fig.~\ref{fig:universe_affinity}. 

In general, all the methods with $\mbfk^{u}$ outperform those with the raw pairwise affinity $\mbfk$. For the recent learning-free solver ZAC, the version of universe affinity has 2\% improvements, and it is much better on GAGM and RRWM (17\% and 14\% respectively). Meanwhile, all the methods tend to obtain similar results when adopting universe affinity, and the red bars of GAGM, RRWM, and ZAC are almost of the same height. It might attribute to the construction of node-wise universe affinity, which simplifies the information of edge affinity and degrades the difficulty of finding the optimal solution.

\begin{figure*}[tb!]
    \centering
    \includegraphics[width=\textwidth]{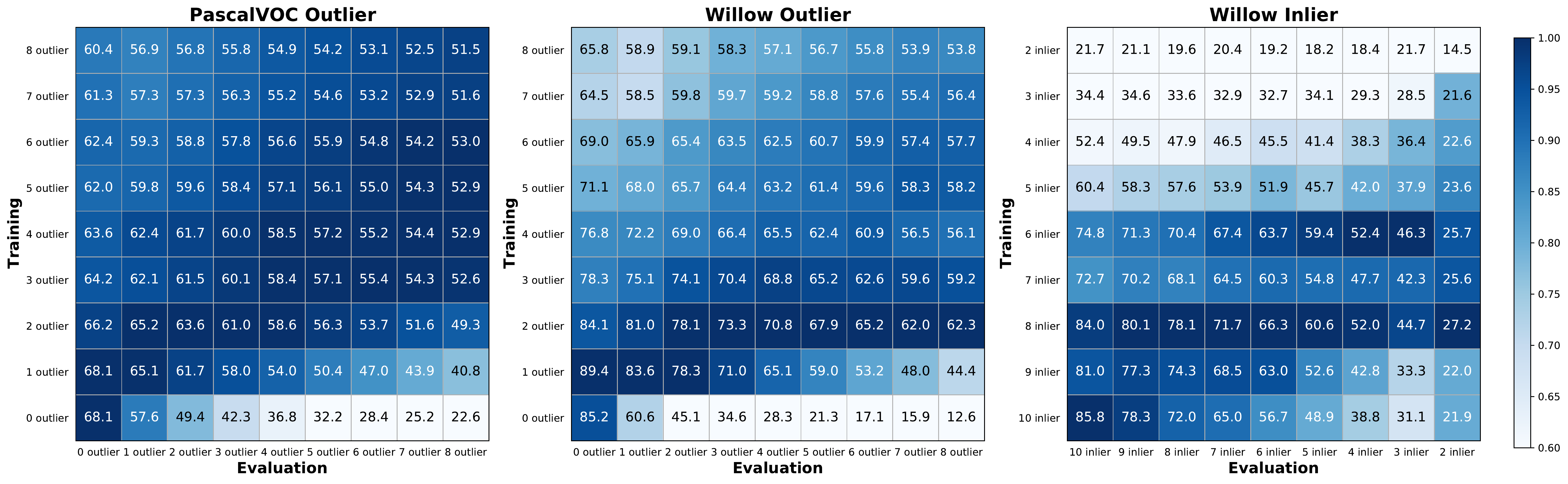}
    \caption{Generalization study on outliers and unmatched inliers shown as confusion matrix of our proposed \our \ on Pascal VOC and Willow Object. Models are trained on the setting of the y-axis and tested on the setting of the x-axis. F1 scores are reported in the entries of confusion matrices. The color map is determined by the F1 score value in the cell normalized by the highest value in its column. Darker color denotes relatively better performance in the same testing setting. The darker the row is, the better the generalization ability of the model.}
    \label{fig:confusion_matrix}
\end{figure*}

\begin{figure*}[tb!]
    \centering
    \includegraphics[width=\textwidth]{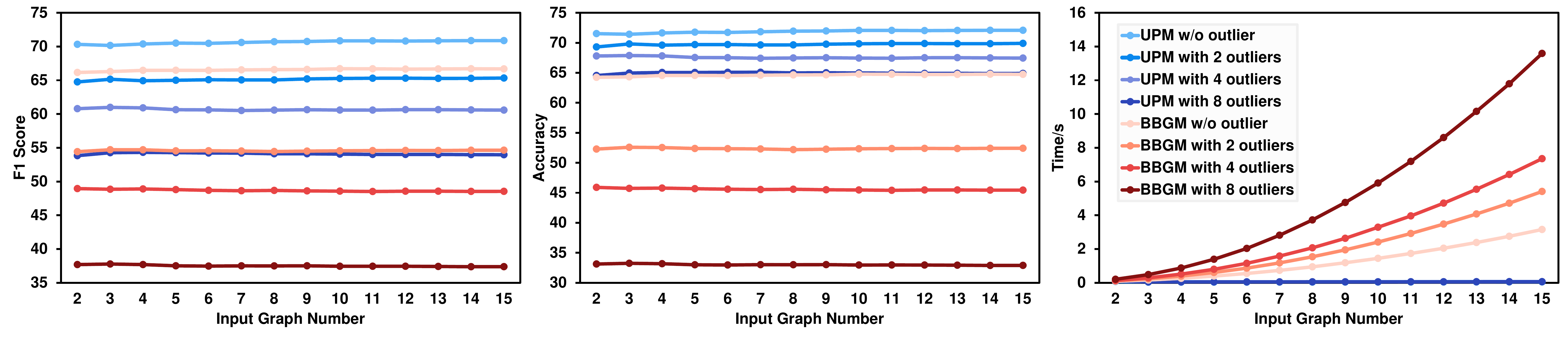}
    %\vspace{-10pt}
    \caption{F1 score, accuracy and time cost on online partial matching. We compare \our\ with BBGM on PascalVOC with two outliers or not.}
    \label{fig:online}
  %  %\vspace{-10pt}
\end{figure*}
\subsubsection{Generalization on unmatched inliers and outliers}
We also test generalization ability for our proposed \our \ on standard partial matching. Here the generalization ability refers to the extent of performance degradation when the evaluation setting is different from training e.g. with different number of outliers or varying number of unmatched inliers. For PascalVOC, we keep all the unmatched inliers and test the generalization ability with 0 to 8 outliers. For Willow Object, we test generalization ability for both unmatched inlier and outlier. We first randomly drop 2 inliers and test the generalization ability with 0 to 8 outliers. Then we randomly add 2 outliers and test the generalization ability with 0 to 8 inliers drop, which means there are 10 to 2 inliers in each graph. The F1 scores are shown in Fig.~\ref{fig:confusion_matrix}. The color map is determined by the F1 score in the current cell normalized by the highest in its column. Darker color denotes relatively better performance in the same testing setting. The darker of the row, the better the generalization ability.

As one can see, \our \ shows a strong generalization ability against the various number of outliers. The model trained with few outliers can achieve the highest F1 score with numerous outliers. The model trained with 3 outliers even achieves the best F1 score when it is tested in the settings with 4/5/6/7 outliers. The model trained with massive outliers can also obtain a good solution under the simple setting. Most of the models exceed 80\% of the highest F1 score, and so does for the case in the middle matrix in Willow Object, where the ratio of inliers and outliers is more ill than PascalVOC. It is interesting to observe that only the model trained without outlier fails to generalize to other cases of massive outliers, which suggests that introducing outliers may improve the generalization ability. 

On the other side, the model trained with few dropped inliers is also able to achieve the highest F1 score in numerous dropped inliers setting, as shown by the model trained with 2 drops. However, when the training cases are too ill-posed, the model's capability dramatically degrades, e.g. there are 10 key points in each graph of Willow Object and we randomly drop 8 inliers in experiments, which means there hardly exists a matched node pair in training or evaluation instance.

Moreover, we do not conduct the confusion matrices over categories. Since we can hardly construct a universe graph for an unknown category, our model \our \ is bound to fail on categories confusion experiments.

\begin{table*}[tb!]
    \caption{Hyper-parameter sensitivity exploration for feature-merged (top) and node-merged (bottom) universe graph respectively. F1 score are reported above. Feature-merged universe graph based \our \ are trained and evaluated on pairs with the same category, while node-merged universe graph based \our \ are trained and evaluated on pairs where half of them are of different categories, e.g. `bike' class get half bikes pairs and half pairs of bike and another category. The ground truth matching between different categories is set to zero matrix. \our-24 and \our-244 show the performance of our method with the exactly universe size (inliers and an absorbing node) under feature-merged and the node-merged universe.}
    \label{tab:voc_usize}
    %\vspace{-5pt}
    \resizebox{\textwidth}{!}{
    \renewcommand\arraystretch{1.4}
        \begin{tabular}{|c|cccccccccccccccccccc|c|}
        \hline
        \textbf{PascalVOC} & \textbf{aero} & \textbf{bike} & \textbf{bird} & \textbf{boat} & \textbf{bottle} & \textbf{bus}  & \textbf{car}  & \textbf{cat}  & \textbf{chair} & \textbf{cow}  & \textbf{table} & \textbf{dog}  & \textbf{horse} & \textbf{motor} & \textbf{person} & \textbf{plant} & \textbf{sheep} & \textbf{sofa} & \textbf{train} & \textbf{tv}   & \textbf{avg}   \\ \hline
        \our-\textbf{15}        & 44.2          & 65.7          & 63.5          & 42.8          & 80.3            & 69.4          & 43.3          & 58.1          & 50.1           & 60.0          & 48.0           & 57.7          & 57.1           & 61.7           & 28.6            & \textbf{86.7}  & 55.5           & 37.3          & 64.4           & 78.1          & 57.63          \\
        \our-\textbf{20}        & 46.0          & 66.6          & 62.6          & 38.7          & \textbf{82.3}   & 71.8          & 50.0          & 70.1          & \textbf{54.7}  & 69.3          & 42.3           & 69.2          & \textbf{66.9}  & 61.8           & 44.5            & \textbf{86.7}  & 61.2           & 45.6          & \textbf{68.8}  & 77.4          & 61.83          \\
        \our-\textbf{24}        & \textbf{47.1} & 64.7          & 63.9          & 43.6          & 78.9            & 70.6          & \textbf{67.1} & 68.5          & 51.7           & 67.9          & 35.7           & 69.3          & 63.5           & 59.1           & 54.7            & 86.4           & \textbf{62.6}  & 36.4          & 65.3           & 78.8          & 61.78          \\
        \our-\textbf{25}        & 46.6          & 65.8          & 64.7          & 39.8          & 80.6            & 70.0          & 52.7          & 69.3          & 54.0           & 69.5          & 55.9           & 69.0          & 63.9           & 62.0           & 51.3            & 85.9           & 59.2           & \textbf{48.1} & 66.6           & 80.4          & 62.75          \\
        \our-\textbf{30}        & 46.3          & \textbf{67.2} & \textbf{65.3} & \textbf{44.9} & 80.5            & \textbf{72.7} & 63.5          & \textbf{71.9} & 49.1           & \textbf{73.5} & \textbf{58.8}  & \textbf{69.4} & 63.6           & 62.4           & \textbf{55.5}   & 86.4           & 63.0           & 28.4          & 68.5           & \textbf{80.8} & \textbf{63.59} \\
        \our-\textbf{35}        & 46.1          & 66.6          & 64.3          & 38.4          & 80.0            & 71.1          & 50            & 70.6          & 50.4           & 69.7          & 31.4           & 68.8          & 63.1           & \textbf{63.7}  & 52.0            & 87.0           & 62.2           & 40.4          & 66.9           & 78.3          & 61.04          \\ \hline
        \our-\textbf{200}       & 43.4          & 63.3          & 56.8          & 40.4          & 78.7            & 64.4          & 47.5          & 64.9          & 49.1           & 64.2          & 42.6           & 66.3          & 56.4           & 58.9           & 36.9            & 78.5           & 58.2           & 36.9          & 55.4           & 70.3          & 56.65          \\
        \our-\textbf{244}       & 43.4          & 57.1          & 54.8          & 35.9          & 78.3            & 69.1          & 53            & 66.1          & 42.3           & 62.1          & 40.2           & 65.9          & 56.6           & 56.5           & 34.3            & \textbf{84.3}  & \textbf{61.0}  & 40.9          & \textbf{64.0}  & \textbf{77.3} & 57.14          \\
        \our-\textbf{300}       & 47.6          & \textbf{66.7} & \textbf{61.6} & \textbf{46.1} & 80.6            & \textbf{69.5} & \textbf{59.2} & \textbf{67.2} & \textbf{55.9}  & \textbf{70.4} & 31.2           & \textbf{70.1} & 64.0           & 63.3           & \textbf{53.3}   & 81.9           & \textbf{61.0}  & 39.6          & 60.1           & 74.9          & \textbf{61.2}  \\
        \our-\textbf{400}       & \textbf{47.7} & 65.8          & 59.9          & 41.8          & 79.3            & 70.0          & 57.0          & 65.9          & 49.5           & 67.5          & \textbf{47.1}  & 68.1          & \textbf{65.5}  & \textbf{63.4}  & 45.2            & 79.5           & 60.3           & \textbf{44.3} & 61.5           & 71.7          & 60.55          \\
        \our-\textbf{500}       & 45.4          & 62.5          & 55.5          & 43.4          & \textbf{80.8}   & 67.4          & 51.7          & 63.0          & 51.9           & 61.6          & 37.0           & 65.5          & 60.8           & 58.7           & 29.2            & 77.1           & 55.9           & 38.6          & 58.7           & 72.74         & 56.9           \\ \hline
        \end{tabular}}
    %%\vspace{-10pt}
\end{table*}

\begin{table*}[tb!]
    \centering
    \caption{Mixture graph matching and clustering on PascalVOC with inference time overhead in seconds. All the unmatched inliers are reserved and two outliers are randomly added for each image.`3$\times$8' denotes that we randomly pick three categories in PascalVOC and for each category, we randomly sample 8 graphs to build the evaluation instance. Instead of MAC, we report the F1C score, which is more suitable for partial matching.}
    \label{tab:voc_clustering}
    %\vspace{-5pt}
    \resizebox{\textwidth}{!}{
    \renewcommand\arraystretch{1.4}
    \begin{tabular}{|c|ccccc|ccccc|ccccc|}
    \hline
    \textbf{PascalVOC}        & \multicolumn{5}{c|}{\textbf{cluster \# $\times$ cluster size: 3$\times$8}}                                                  & \multicolumn{5}{c|}{\textbf{cluster \# $\times$ cluster size: 4$\times$6}}                                                 & \multicolumn{5}{c|}{\textbf{cluster \# $\times$ cluster size: 6$\times$4}}                                                 \\ \cline{2-16} 
                       & \textbf{F1C}$\uparrow$    & \textbf{CP}$\uparrow$    & \textbf{RI}$\uparrow$    & \textbf{CA}$\uparrow$    & \textbf{time}$\downarrow$ & \textbf{F1C}$\uparrow$    & \textbf{CP}$\uparrow$    & \textbf{RI}$\uparrow$    & \textbf{CA}$\uparrow$    & \textbf{time}$\downarrow$ & \textbf{F1C}$\uparrow$    & \textbf{CP}$\uparrow$    & \textbf{RI}$\uparrow$    & \textbf{CA}$\uparrow$    & \textbf{time}$\downarrow$ \\ \hline
    \textbf{BBGM~\cite{rolinek2020deep}}    & 52.68          & 83.89          & 85.51          & 81.75          & 2.36          & 52.37          & 84.24          & 86.46          & 82.06          & 2.36          & 53.25          & \textbf{80.70}          & 89.42          & \textbf{79.08}          & 3.48          \\
    \textbf{UPM-30}  & 50.16          & 81.15          & 81.59          & 77.01          & \textbf{0.32} & 47.33          & 78.02          & 83.72          & 72.74          & \textbf{0.33}          & 48.69          & 71.62          & 84.86          & 67.90          & \textbf{0.43}          \\
    \textbf{UPM-300} & \textbf{64.79} & \textbf{89.69} & \textbf{89.94} & \textbf{87.09} & 0.34          & \textbf{63.78} & \textbf{87.67} & \textbf{90.91} & \textbf{84.62} & \textbf{0.33} & \textbf{64.12} & 80.33 & \textbf{90.45} & 77.00 & \textbf{0.43} \\ \hline
    \end{tabular}}
\end{table*}

\begin{table}[tb!]
    \centering
    \caption{Mixture graph matching and clustering on Willow Object with inference time overhead in seconds. We follow \cite{wang2020graduated} by randomly picking 8 cars, 8 ducks and 8 motorbike without outliers and unmatched inliers.}
    %\vspace{-5pt}
    \renewcommand\arraystretch{1.2}
    \begin{tabular}{|c|ccccc|}
    \hline
    \textbf{Willow} & \textbf{CP}$\uparrow$ & \textbf{RI}$\uparrow$ & \textbf{CA}$\uparrow$ & \textbf{MAC}$\uparrow$ & \textbf{time}$\downarrow$ \\ \hline
    \textbf{RRWM~\cite{Cho2010ReweightedRW}}      & 87.9       & 87.1       & 81.5       & 74.8       & \textbf{0.4}           \\
    \textbf{CAO-c~\cite{YanPAMI16}}     & 90.8       & 90.3       & 86.0       & 87.8       & 3.3           \\
    \textbf{CAO-pc~\cite{YanPAMI16}}    & 88.7       & 88.3       & 83.1       & 87.0       & 1.8           \\
    \textbf{DPMC~\cite{WangAAAI20}}      & 93.1       & 92.3       & 89.0       & 87.2       & 1.2           \\
    \textbf{GA-MGMC~\cite{wang2020graduated}}   & 92.1       & 90.5       & 89.3       & 65.3       & 10.6          \\
    \textbf{GANN-MGMC~\cite{wang2020graduated}} & 97.6       & 97.0       & 96.3       & 89.6       & 5.2           \\ \hline
    \our-\textbf{90}    & \textbf{98.92}      & \textbf{98.59}      & \textbf{98.18}      & \textbf{94.24}      & 0.52          \\ \hline
    \end{tabular}
    \label{tab:willow_clustering}
\end{table}

\subsubsection{Hyper-parameter sensitivity}
We conduct a sensitivity test for universe graph size $n_u$ via two schemes of construction. For the feature-merged universe graph, we apply the standard partial matching setting on PascalVOC with unmatched inliers and two outliers. Since $n_u$ is defined as node feature number of sub-universe graphs $n_u = \max n_i$, we vary the $n_u$ from 15 to 35 to see its performance. For the node-merged universe graph, we apply the mixture partial matching setting mentioned above, where half of both training and test sample pairs are of different categories. Unmatched inliers and two outliers are adopted as well. We report  F1 score of \our \ with $n_u = 200,300,400,500$. Note that the category of person has the most kinds of labels: 23 different kinds of labels\footnote{For example, the category `person' have different kinds of labels for key points: `left eye', `right eye', `left hand', `right hand', and so on.}, whereas other categories have 6 $\sim$ 16. The sum of label kinds over all categories is 243. (\our-24 and \our-244 add a absorbing node)

As shown in Table~\ref{tab:voc_usize}, our method \our \ is robust against universe graph size. As for feature-merged universe graph, even 40\% overestimate or underestimate of $n_u$ only results in 6\% performance drop. On the other side, there is only 4.3\% decrease on F1 score with doubled $n_u$ (\our-\textbf{500} for $n_u=244$) in node-merged universe graph. Moreover, we also find that the impact of underestimating is more serious than the impact of overestimating. \our \ even encourages a little bit overestimating to achieve the best performance. As shown in Table~\ref{tab:voc_usize}, \our-\textbf{30} outperforms \our-\textbf{24} with 1.81\% on F1 score, where the latter is a better estimation for the exact universe size $n_u=23$.

\subsection{Extension to Online and Mixtured Matching} \label{subsec:extension_case}
In this subsection, we show our model \our's performance in two extension cases: online multiple graph matching~\cite{YuECCV18,ChenECCV20}, and mixture categories matching and clustering~\cite{WangAAAI20}. 

The former relates to matching new coming graphs given the computed matchings among existing graphs. While the latter refers to the setting of solving graph clustering and matching within the clusters, which is a more challenging task than traditional graph matching that assumes all the graphs are matchable within the same category.

\begin{figure*}[tb!]
    \centering
    \includegraphics[width=1.0\textwidth]{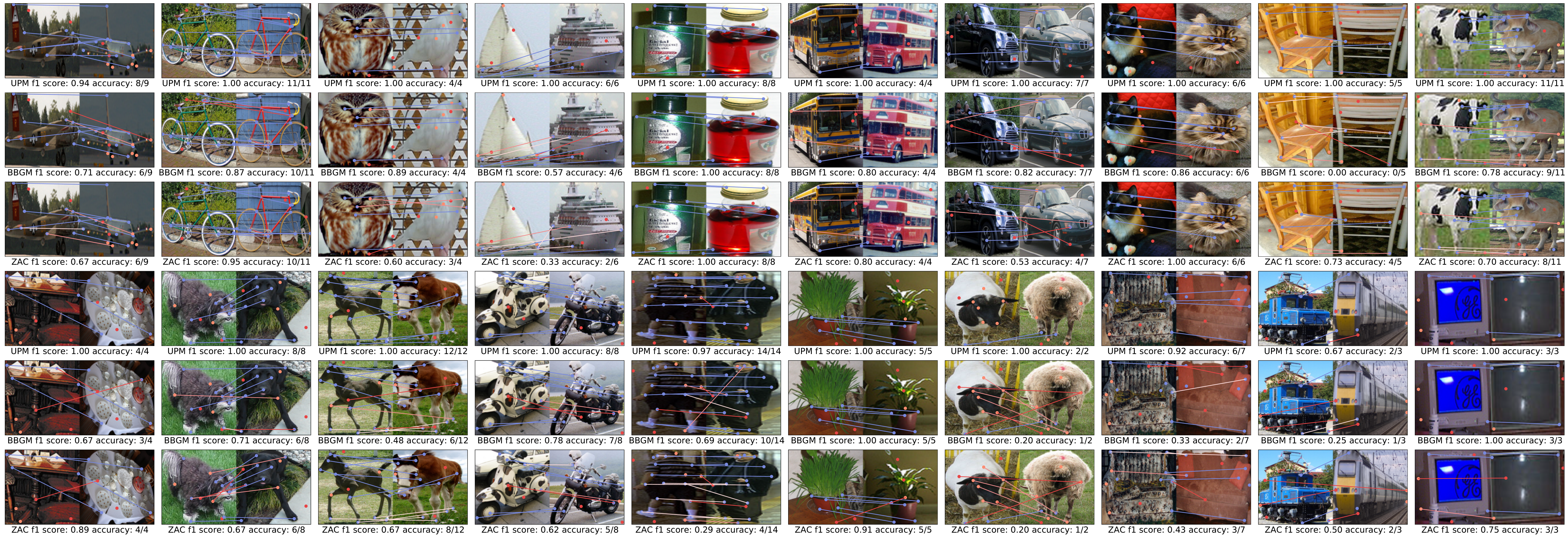}
    \caption{Visualization of matching results on 20 Pascal VOC Keypoint categories. The legend follows Fig.~\ref{fig:node_type}. Blue, orange, and red nodes denote matched inlier, unmatched inlier, and outlier. Blue lines represent correct and three types of red lines represent mismatching, ill-matching, and over-matching respectively. \our \, BBGM, and ZAC are eval at the same matching pair, where F1 score and accuracy are reported at the bottom.}
    \label{fig:visualization_cmp}
\end{figure*}

\begin{figure*}[tb!]
    \centering
    \includegraphics[width=1.0\textwidth]{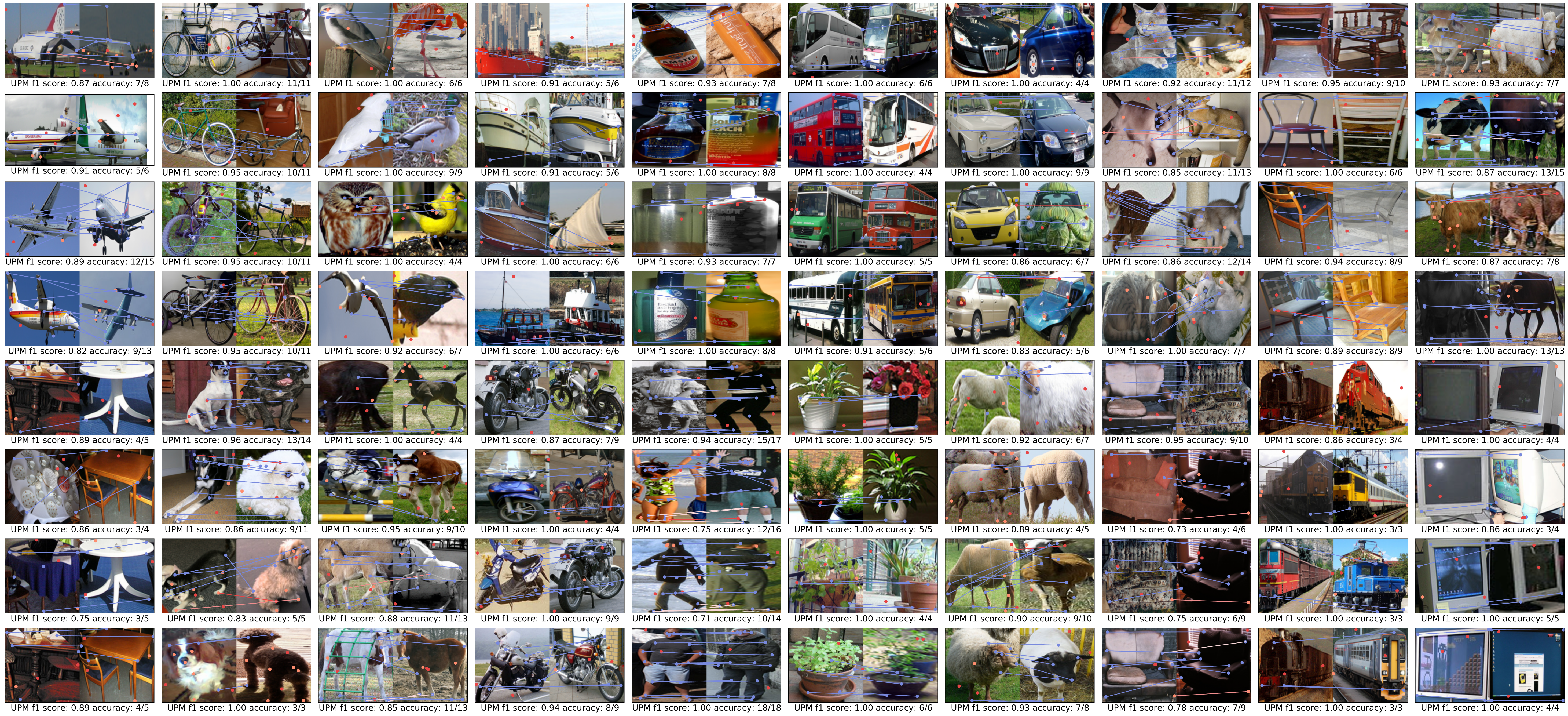}
    \caption{Visualization of \our \ matching results on 20 Pascal VOC Keypoint categories. The legend follows Fig.~\ref{fig:node_type}. Blue, orange, and red nodes denote matched inlier, unmatched inlier, and outlier. Blue lines represent correct and three types of red lines represent mismatching, ill-matching, and over-matching respectively. We show the success of \our \ against rotation, occlusion, and massive inliers.}
    \label{fig:visualization_show}
\end{figure*}

\subsubsection{Online multiple graph matching}
We randomly pick 15 graphs of the same category in PascalVOC. We do not filter out any unmatched inlier and randomly add 0/2/4/8 outliers for each graph. Then the graphs are input to the model one by one. For BBGM, we match the new coming graph with all existing graphs. For \our, we directly match the new coming graph to the universe graph and compute the matching from the new graph to the others by $\hat{\mbfx}_{ij}=\hat{\mbfx}_{iu} \hat{\mbfx}_{ju}^{\top}$. As one can see, each coming graph will only be matched to the universe graph once, and thus the time cost is a constant term with the number of graphs increasing. We evaluate the F1 score, accuracy as well as time cost, and the average results over all the categories are shown in Fig.~\ref{fig:online}.

The F1 score and accuracy keep almost unchanged every time a new graph arrives for all the methods. However, our method \our \ runs faster than BBGM, where the time cost of BBGM increases linearly or even faster, while the time cost of \our \ almost stays the same as the number of graphs grows. This reveals the advantage of our universe matching mechanism on accelerating the speed of the solver.

\subsubsection{Mixture graph matching and clustering}
We conduct two experiments for mixture graph matching and clustering. All the learning-based models e.g. \our \ and BBGM are trained on Willow or PascalVOC where half of training sample pairs are of the same category and half are of different categories. During the evaluation, the clustering results are obtained via spectral clustering~\cite{ng2002spectral} based on affinity score, which follows the process of DPMC~\cite{WangAAAI20} and GA-MGMC~\cite{wang2020graduated}.

We first compare our method with learning-free solvers on Willow Object. Evaluation is performed with 8 Cars, 8 Ducks, and 8 Motorbikes, and two outliers are randomly added for each graph. The results are shown in Table~\ref{tab:willow_clustering} where the performance of other methods are reported according to the number in \cite{wang2020graduated}. As one can see, \our \ outperforms other methods on all the clustering metrics, with a low cost of inference time. It is 2$\times \sim$  20$\times$ faster than solver that achieves good performance. While compare with the RRWM with the same magnitude time cost, \our \ outperforms over 10\% on all the matching quality evaluation metrics. 

However, the learning-based models all well perform on Willow-Object which is less challenging. While on PascalVOC, which is much more complicated than Willow-Object due to the occlusion and rotation perspective, we compare our method \our-\textbf{30} (feature-merged universe graph) and \our-\textbf{300} (node-merged universe graph) with BBGM as the baseline. During the evaluation phase, we resample the cluster as well as graphs for each inference. Different sampling strategy e.g. `3$\times$8', `4$\times$6', and `6$\times$4' are applied, where `3$\times$8' denotes that we randomly pick three categories in PascalVOC and for each category, we random sample 8 graphs to build the evaluation instance.

As shown in Table~\ref{tab:voc_clustering}, \our-\textbf{300} outperforms the other two methods in terms of both cluster quality and speed. In both of the 3$\times$8 and 4$\times$6 settings, \our-\textbf{300} achieves an improvement of 3\%-6\% on clustering metrics as \textbf{CP}, \textbf{RI}, and \textbf{CA}, while matching quality \textbf{F1C} and speed enjoy a remarkable improvement. 
The gap on cluster metrics e.g. CA, RI, and CP decrease by raising the number of image categories. However \our-\textbf{300} still has a great advantage on F1 score. 

On the other hand, \our-\textbf{30} fails to cope with mixture graph matching and clustering.
Feature-merged universe graph mixes up the nodes from all the sub-universe graphs, and thus fails to reject the mapping over classes 
It also supports our hypothesis on the structure of the feature-merged graph and the node-merged universe graph.

\section{Conclusion and Outlook}
In this paper, we first introduce the universe graph into deep graph matching model. Universe graph offers a matching template that gathers the nodes of the same type with anchor points and optimizes the node feature distribution in feature space. Based on that, we propose our method \textbf{U}niverse \textbf{P}artial \textbf{M}atching (\our). Benefit from the mechanism of universe graph, \our \ serves as a unified model for all the tasks in graph matching and is able to distinguish node type at a fine-grained level. It achieves the best performance in all the tasks under the partial matching setting and is robust against the variance of outliers and inliers. 

Promising results as \our \ shows, it still relies on a large number of labeled pairs. Such corresponding label is expensive and time-consuming especially for some specific fields e.g. drug discovery and video tracing. Unsupervised models for graph matching are still under exploration. Meanwhile, \our \ has high precision and a low recall on matching prediction. It can be further improved in future works to balance them or adjust the ratio with a hyperparameter.

% In this paper, we first introduce the multiple graph matching perspective and universe matching into the deep learning of graph matching. It broadens the view of the training pair, and fuses the information carried by all the graphs in dataset to build a universe graph. Based on that, we propose our method \textbf{U}niverse \textbf{P}artial \textbf{M}atching (\our), which achieves the best performance in all the partial matching setting, and is robust against the variance of outliers and inliers. Moreover, it has the ability to deal with pairwise matching as well as multiple graph matching, and thus can be applied to both online matching and mixture graph matching and clustering. 

\section*{Acknowledgement}
This work was partly supported by National Key Research and Development Program of China (2020AAA0107600), Shanghai Municipal Science and Technology Major Project (2021SHZDZX0102), and NSFC (61972250, 72061127003). The authors are also thankful to the valuable comments from the reviewers and associate editor.

\bibliographystyle{IEEEtran}
\bibliography{main}

% Generated by IEEEtran.bst, version: 1.14 (2015/08/26)
\begin{thebibliography}{10}
\providecommand{\url}[1]{#1}
\csname url@samestyle\endcsname
\providecommand{\newblock}{\relax}
\providecommand{\bibinfo}[2]{#2}
\providecommand{\BIBentrySTDinterwordspacing}{\spaceskip=0pt\relax}
\providecommand{\BIBentryALTinterwordstretchfactor}{4}
\providecommand{\BIBentryALTinterwordspacing}{\spaceskip=\fontdimen2\font plus
\BIBentryALTinterwordstretchfactor\fontdimen3\font minus
  \fontdimen4\font\relax}
\providecommand{\BIBforeignlanguage}[2]{{%
\expandafter\ifx\csname l@#1\endcsname\relax
\typeout{** WARNING: IEEEtran.bst: No hyphenation pattern has been}%
\typeout{** loaded for the language `#1'. Using the pattern for}%
\typeout{** the default language instead.}%
\else
\language=\csname l@#1\endcsname
\fi
#2}}
\providecommand{\BIBdecl}{\relax}
\BIBdecl

\bibitem{ShenTMI02}
D.~Shen and C.~D. Hammer, ``Hierarchical attribute matching mechanism for
  elastic registration,'' \emph{TMI}, 2002.

\bibitem{bregler2000recovering}
C.~Bregler, A.~Hertzmann, and H.~Biermann, ``Recovering non-rigid 3d shape from
  image streams,'' in \emph{CVPR}, vol.~2, 2000, pp. 690--696.

\bibitem{vijayanarasimhan2017sfm}
S.~Vijayanarasimhan, S.~Ricco, C.~Schmid, R.~Sukthankar, and K.~Fragkiadaki,
  ``Sfm-net: Learning of structure and motion from video,'' \emph{arXiv
  preprint arXiv:1704.07804}, 2017.

\bibitem{nam2016learning}
H.~Nam and B.~Han, ``Learning multi-domain convolutional neural networks for
  visual tracking,'' in \emph{CVPR}, 2016, pp. 4293--4302.

\bibitem{iqbal2017posetrack}
U.~Iqbal, A.~Milan, and J.~Gall, ``Posetrack: Joint multi-person pose
  estimation and tracking,'' in \emph{CVPR}, 2017, pp. 2011--2020.

\bibitem{baker2011database}
S.~Baker, D.~Scharstein, J.~Lewis, S.~Roth, M.~J. Black, and R.~Szeliski, ``A
  database and evaluation methodology for optical flow,'' \emph{IJCV}, vol.~92,
  no.~1, pp. 1--31, 2011.

\bibitem{sun2014quantitative}
D.~Sun, S.~Roth, and M.~J. Black, ``A quantitative analysis of current
  practices in optical flow estimation and the principles behind them,''
  \emph{IJCV}, vol. 106, no.~2, pp. 115--137, 2014.

\bibitem{sun2018pwc}
D.~Sun, X.~Yang, M.-Y. Liu, and J.~Kautz, ``Pwc-net: Cnns for optical flow
  using pyramid, warping, and cost volume,'' in \emph{CVPR}, 2018, pp.
  8934--8943.

\bibitem{luo2016efficient}
W.~Luo, A.~G. Schwing, and R.~Urtasun, ``Efficient deep learning for stereo
  matching,'' in \emph{CVPR}, 2016, pp. 5695--5703.

\bibitem{chang2018pyramid}
J.-R. Chang and Y.-S. Chen, ``Pyramid stereo matching network,'' in
  \emph{CVPR}, 2018, pp. 5410--5418.

\bibitem{cao2017realtime}
Z.~Cao, T.~Simon, S.-E. Wei, and Y.~Sheikh, ``Realtime multi-person 2d pose
  estimation using part affinity fields,'' in \emph{CVPR}, 2017, pp.
  7291--7299.

\bibitem{gasse2019exact}
M.~Gasse, D.~Ch{\'e}telat, N.~Ferroni, L.~Charlin, and A.~Lodi, ``Exact
  combinatorial optimization with graph convolutional neural networks,''
  \emph{arXiv preprint arXiv:1906.01629}, 2019.

\bibitem{Gold1996AGA}
S.~Gold and A.~Rangarajan, ``A graduated assignment algorithm for graph
  matching,'' \emph{IEEE TPAMI}, vol.~18, pp. 377--388, 1996.

\bibitem{Cho2010ReweightedRW}
M.~Cho, J.~Lee, and K.~M. Lee, ``Reweighted random walks for graph matching,''
  in \emph{ECCV}, 2010, pp. 492--505.

\bibitem{lee2011hyper}
J.~Lee, M.~Cho, and K.~M. Lee, ``Hyper-graph matching via reweighted random
  walks,'' in \emph{CVPR}, 2011, pp. 1633--1640.

\bibitem{yan2015consistency}
J.~Yan, J.~Wang, H.~Zha, X.~Yang, and S.~Chu, ``Consistency-driven alternating
  optimization for multigraph matching: A unified approach,'' \emph{IEEE TIP},
  vol.~24, no.~3, pp. 994--1009, 2015.

\bibitem{ZhouICCV15}
X.~Zhou, M.~Zhu, and K.~Daniilidis, ``Multi-image matching via fast alternating
  minimization,'' in \emph{ICCV}, 2015, pp. 4032--4040.

\bibitem{YanPAMI16}
J.~Yan, M.~Cho, H.~Zha, X.~Yang, and S.~M. Chu, ``Multi-graph matching via
  affinity optimization with graduated consistency regularization,'' \emph{IEEE
  TPAMI}, vol.~38, no.~6, pp. 1228--1242, 2015.

\bibitem{JiangPAMI21}
Z.~{Jiang}, T.~{Wang}, and J.~{Yan}, ``Unifying offline and online multi-graph
  matching via finding shortest paths on supergraph,'' \emph{IEEE TPAMI},
  vol.~43, no.~10, pp. 3648--3663, 2021.

\bibitem{YanICMR16}
J.~Yan, X.-C. Yin, W.~Lin, C.~Deng, H.~Zha, and X.~Yang, ``A short survey of
  recent advances in graph matching,'' in \emph{ICMR}, 2016.

\bibitem{YanIJCAI20}
J.~Yan, S.~Yang, and E.~Hancock, ``Learning graph matching and related
  combinatorial optimization problems,'' in \emph{IJCAI}, 2020.

\bibitem{ZanfirCVPR18}
A.~Zanfir and C.~Sminchisescu, ``Deep learning of graph matching,'' in
  \emph{CVPR}, 2018, pp. 2684--2693.

\bibitem{WangICCV19}
R.~Wang, J.~Yan, and X.~Yang, ``Learning combinatorial embedding networks for
  deep graph matching,'' in \emph{ICCV}, 2019, pp. 3056--3065.

\bibitem{wang2020learning}
T.~Wang, H.~Liu, Y.~Li, Y.~Jin, X.~Hou, and H.~Ling, ``Learning combinatorial
  solver for graph matching,'' in \emph{CVPR}, 2020, pp. 7568--7577.

\bibitem{Wang2019Neural}
R.~Wang, J.~Yan, and X.~Yang, ``Neural graph matching network: Learning
  lawler’s quadratic assignment problem with extension to hypergraph and
  multiple-graph matching,'' \emph{IEEE TPAMI}, 2021.

\bibitem{rolinek2020deep}
M.~Rol{\'\i}nek, P.~Swoboda, D.~Zietlow, A.~Paulus, V.~Musil, and G.~Martius,
  ``Deep graph matching via blackbox differentiation of combinatorial
  solvers,'' in \emph{ECCV}, 2020, pp. 407--424.

\bibitem{yu2021deep}
T.~Yu, R.~Wang, J.~Yan, and B.~Li, ``Deep latent graph matching,'' in
  \emph{ICML}, 2021, pp. 12\,187--12\,197.

\bibitem{WangAAAI20}
T.~Wang, Z.~Jiang, and J.~Yan, ``Clustering-aware multiple graph matching via
  decayed pairwise matching composition,'' \emph{AAAI}, 2020.

\bibitem{chen2014near}
Y.~Chen, L.~Guibas, and Q.~Huang, ``Near-optimal joint object matching via
  convex relaxation,'' in \emph{ICML}, 2014, pp. 100--108.

\bibitem{PachauriNIPS13}
D.~Pachauri, R.~Kondor, and V.~Singh, ``Solving the multi-way matching problem
  by permutation synchronization,'' in \emph{NeurIPS}, 2013, pp. 1860--1868.

\bibitem{DymTOG17}
N.~Dym, H.~Maron, and Y.~Lipman, ``Ds++ a flexible, scalable and provably tight
  relaxation for matching problems,'' \emph{ACM Transactions on Graphics
  (TOG)}, vol.~36, no.~6, pp. 1--14, 2017.

\bibitem{bernard2018ds}
F.~Bernard, C.~Theobalt, and M.~Moeller, ``Ds*: Tighter lifting-free convex
  relaxations for quadratic matching problems,'' in \emph{CVPR}, 2018, pp.
  4310--4319.

\bibitem{swoboda2017study}
P.~Swoboda, C.~Rother, H.~Abu~Alhaija, D.~Kainmuller, and B.~Savchynskyy, ``A
  study of lagrangean decompositions and dual ascent solvers for graph
  matching,'' in \emph{CVPR}, 2017, pp. 1607--1616.

\bibitem{HuangSGP13}
Q.-X. Huang and L.~Guibas, ``Consistent shape maps via semidefinite
  programming,'' in \emph{Computer Graphics Forum}, vol.~32, no.~5.\hskip 1em
  plus 0.5em minus 0.4em\relax Wiley Online Library, 2013, pp. 177--186.

\bibitem{wang2018multi}
Q.~Wang, X.~Zhou, and K.~Daniilidis, ``Multi-image semantic matching by mining
  consistent features,'' in \emph{CVPR}, 2018, pp. 685--694.

\bibitem{SwobodaCVPR19}
P.~Swoboda, A.~Mokarian, C.~Theobalt, F.~Bernard \emph{et~al.}, ``A convex
  relaxation for multi-graph matching,'' in \emph{CVPR}, 2019, pp.
  11\,156--11\,165.

\bibitem{YuECCV18}
T.~Yu, J.~Yan, W.~Liu, and B.~Li, ``Incremental multi-graph matching via
  diversity and randomness based graph clustering,'' in \emph{ECCV}, 2018, pp.
  139--154.

\bibitem{wang2020graduated}
R.~Wang, J.~Yan, and X.~Yang, ``Graduated assignment for joint multi-graph
  matching and clustering with application to unsupervised graph matching
  network learning.'' in \emph{NeurIPS}, 2020.

\bibitem{WangPAMI20}
R.~{Wang}, J.~{Yan}, and X.~{Yang}, ``Combinatorial learning of robust deep
  graph matching: an embedding based approach,'' \emph{IEEE TPAMI}, 2020.

\bibitem{ZhangICCV19}
Z.~Zhang and W.~S. Lee, ``Deep graphical feature learning for the feature
  matching problem,'' in \emph{ICCV}, 2019, pp. 5087--5096.

\bibitem{AdamsArxiv11}
R.~P. Adams and R.~S. Zemel, ``Ranking via sinkhorn propagation,''
  \emph{arXiv:1106.1925}, 2011.

\bibitem{YuICLR20}
T.~Yu, R.~Wang, J.~Yan, and B.~Li, ``Learning deep graph matching with
  channel-independent embedding and hungarian attention,'' in \emph{ICLR},
  2020.

\bibitem{KuhnNavalResearch55}
H.~W. Kuhn, ``The hungarian method for the assignment problem,'' in
  \emph{Export. Naval Research Logistics Quarterly}, 1955, pp. 83--97.

\bibitem{FeyCVPR18}
M.~Fey, J.~Eric~Lenssen, F.~Weichert, and H.~M{\"u}ller, ``Splinecnn: Fast
  geometric deep learning with continuous b-spline kernels,'' in \emph{CVPR},
  2018, pp. 869--877.

\bibitem{poganvcic2019differentiation}
M.~V. Pogan{\v{c}}i{\'c}, A.~Paulus, V.~Musil, G.~Martius, and M.~Rolinek,
  ``Differentiation of blackbox combinatorial solvers,'' in \emph{ICLR}, 2019.

\bibitem{RolinekECCV20}
M.~Rol{\'\i}nek, P.~Swoboda, D.~Zietlow, A.~Paulus, V.~Musil, and G.~Martius,
  ``Deep graph matching via blackbox differentiation of combinatorial
  solvers,'' in \emph{ECCV}, 2020, pp. 407--424.

\bibitem{ying2020neural}
Z.~Ying, A.~Wang, J.~You, C.~Wen, A.~Canedo, and J.~Leskovec, ``Neural subgraph
  matching,'' \emph{arXiv preprint arXiv:2007.03092}, 2020.

\bibitem{bai2021glsearch}
Y.~Bai, D.~Xu, Y.~Sun, and W.~Wang, ``Glsearch: Maximum common subgraph
  detection via learning to search,'' in \emph{ICML}, 2021, pp. 588--598.

\bibitem{he2016deep}
K.~He, X.~Zhang, S.~Ren, and J.~Sun, ``Deep residual learning for image
  recognition,'' in \emph{CVPR}, 2016, pp. 770--778.

\bibitem{ZhouPAMI16}
F.~Zhou and F.~Torre, ``Factorized graph matching,'' \emph{IEEE TPAMI}, 2016.

\bibitem{ng2002spectral}
A.~Y. Ng, M.~I. Jordan, and Y.~Weiss, ``On spectral clustering: Analysis and an
  algorithm,'' in \emph{NeurIPS}, 2002, pp. 849--856.

\bibitem{Wang2018GraphMW}
T.~Wang, H.~Ling, C.~Lang, and S.~Feng, ``Graph matching with adaptive and
  branching path following,'' \emph{IEEE TPAMI}, 2017.

\bibitem{wang2020zero}
F.~Wang, N.~Xue, J.-G. Yu, and G.-S. Xia, ``Zero-assignment constraint for
  graph matching with outliers,'' in \emph{CVPR}, 2020, pp. 3033--3042.

\bibitem{PascalVOC}
M.~Everingham, L.~Gool, C.~K. Williams, J.~Winn, and A.~Zisserman, ``The pascal
  visual object classes (voc) challenge,'' \emph{IJCV}, vol.~88, no.~2, p.
  303–338, Jun. 2010.

\bibitem{Bourdev2009PoseletsBP}
L.~D. Bourdev and J.~Malik, ``Poselets: Body part detectors trained using 3d
  human pose annotations,'' \emph{ICCV}, pp. 1365--1372, 2009.

\bibitem{Cho2013LearningGT}
M.~Cho, A.~Karteek, and J.~Ponce, ``Learning graphs to match,'' \emph{ICCV},
  pp. 25--32, 2013.

\bibitem{caltech256}
G.~Griffin, A.~Holub, and P.~Perona, ``Caltech-256 object category dataset,''
  California Institute of Technology, Tech. Rep., 2007.

\bibitem{everingham2007pascal}
M.~Everingham, L.~Van~Gool, C.~K.~I. Williams, J.~Winn, and A.~Zisserman, ``The
  {PASCAL} {V}isual {O}bject {C}lasses {C}hallenge 2007 {(VOC2007)}
  {R}esults,''
  http://www.pascal-network.org/challenges/VOC/voc2007/workshop/index.html.

\bibitem{simonyanICLR14vgg}
K.~Simonyan and A.~Zisserman, ``Very deep convolutional networks for
  large-scale image recognition,'' in \emph{ICLR}, 2014.

\bibitem{YanCVPR15}
J.~Yan, C.~Zhang, H.~Zha, W.~Liu, X.~Yang, and S.~Chu, ``Discrete hyper-graph
  matching,'' in \emph{CVPR}, 2015.

\bibitem{TianECCV12}
Y.~Tian, J.~Yan, H.~Zhang, Y.~Zhang, X.~Yang, and H.~Zha, ``On the convergence
  of graph matching: Graduated assignment revisited,'' in \emph{ECCV}, 2012.

\bibitem{ChenECCV20}
Z.~Chen, Z.~Xie, J.~Yan, Y.~Zheng, and X.~Yang, ``Layered neighborhood
  expansion for incremental multiple graph matching,'' in \emph{ECCV}, 2020.

\end{thebibliography}

\begin{IEEEbiography}[{\includegraphics[width=1in,height=1.25in,clip,keepaspectratio]{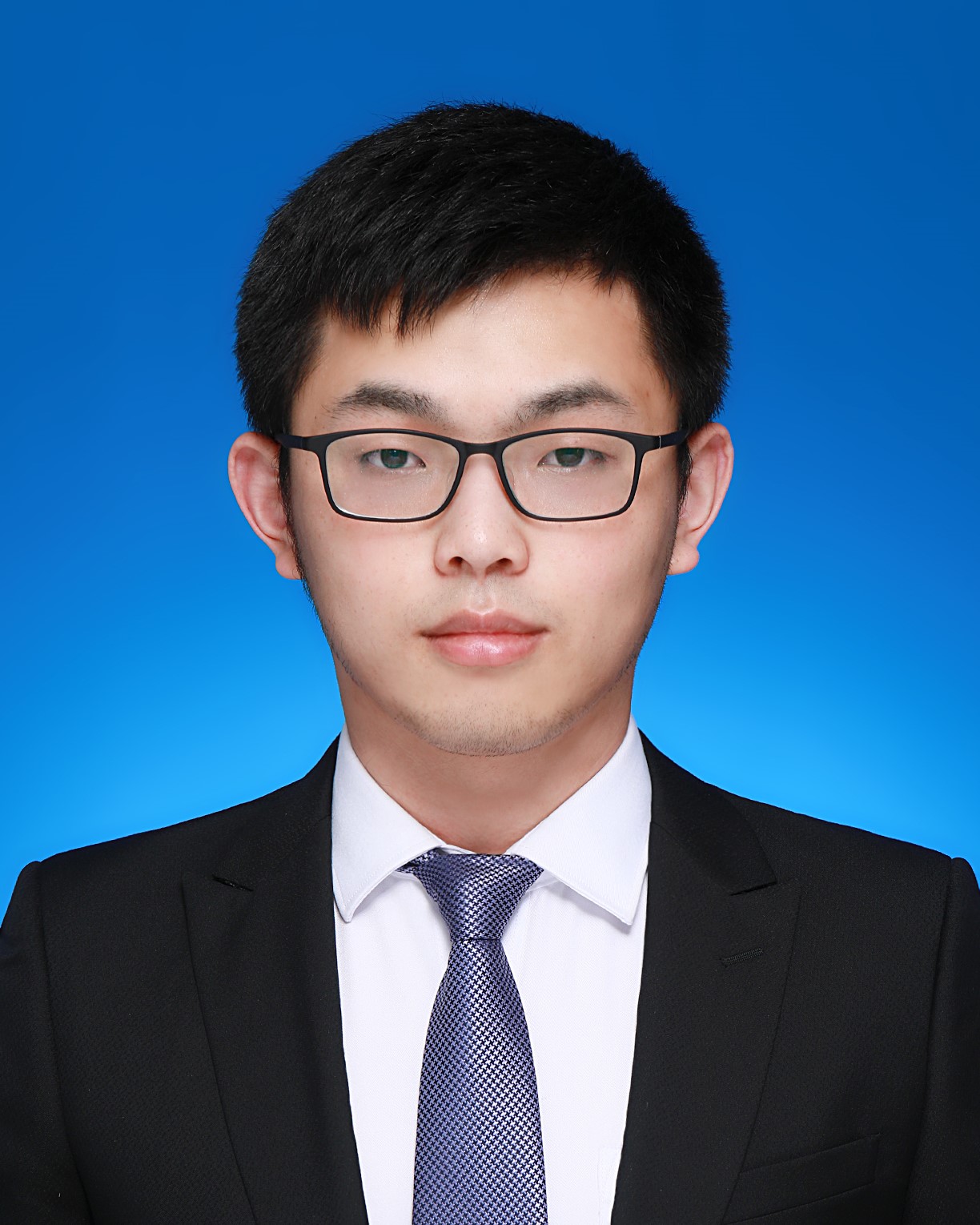}}]{Zetian Jiang} is currently a PhD Student with Department of Computer Science and Engineering, Shanghai Jiao Tong University, Shanghai, China. He received the Bachelor of Engineering in Computer Science and Technology (ACM-Class), Shanghai Jiao Tong University in 2020. His research interests include machine learning and combinatorial optimization. He has published first-authored papers in IEEE TPAMI and AAAI 2020 on combinatorial optimization. 
\end{IEEEbiography}

 \begin{IEEEbiography}[{\includegraphics[width=1in,height=1.25in,clip,keepaspectratio]{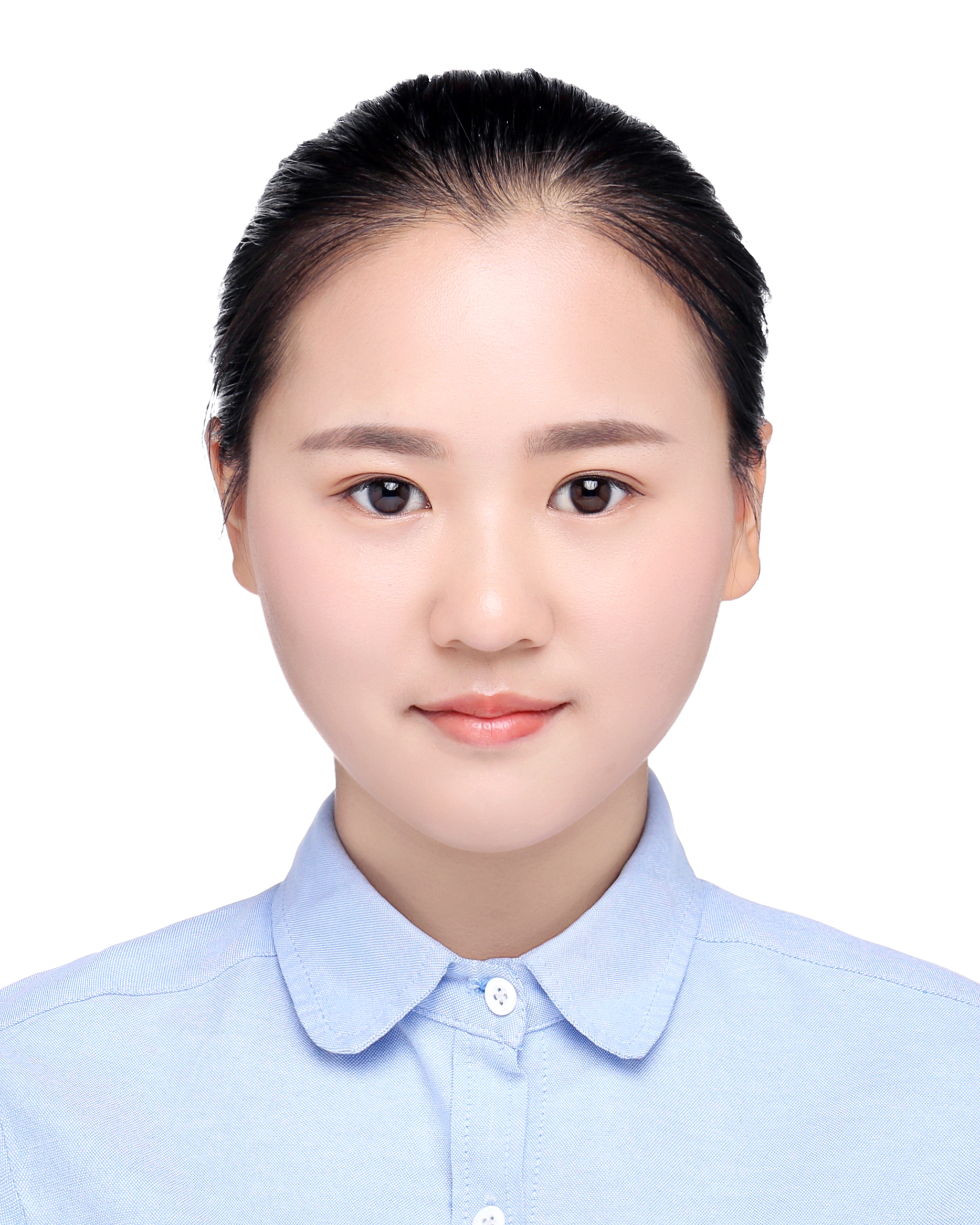}}]{Jiaxin Lu} is currently an Undergraduate Student with ACM Class in Zhiyuan College (Honored Program), Shanghai Jiao Tong University. She once won the third place in Chinese Collegiate Programming Contest Women Final. She worked as research intern with Department of Computer Science, University of Texas at Austin. Her research interests include machine learning and graph analysis.  
 \end{IEEEbiography}
 
 \begin{IEEEbiography}[{\includegraphics[width=1in,height=1.25in,clip,keepaspectratio]{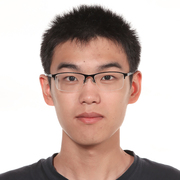}}]{Tianzhe Wang} is currently a Master Student with Department of Computer Science and Engineering, Georgia Institute of Technology, Atlanta, USA. He received the Bachelor of Engineering in Computer Science and Technology (ACM-Class), Shanghai Jiao Tong University in 2020. His research interests include machine learning and combinatorial optimization. He has published first-authored papers in IEEE TPAMI and AAAI20 on combinatorial optimization. 
\end{IEEEbiography}

\begin{IEEEbiography}[{\includegraphics[width=1in,height=1.25in,clip,keepaspectratio]{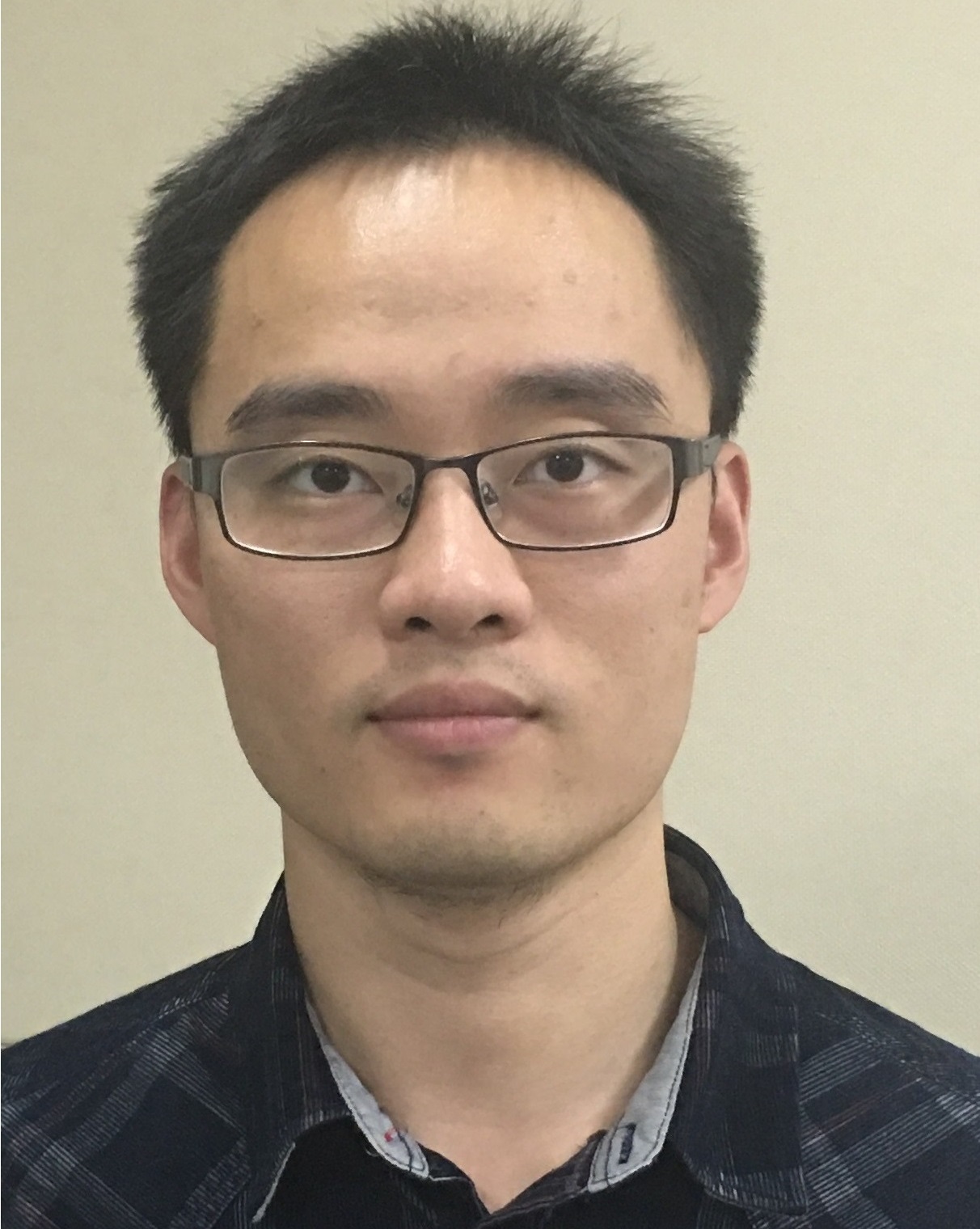}}]{Junchi Yan} (S'10-M'11-SM'21) is an Associate Professor with Shanghai Jiao Tong University, Shanghai, China. Before that, he was a Senior Research Staff Member and Principal Scientist with IBM Research where he started his career since April 2011. He obtained the Ph.D. in Electronic Engineering from Shanghai Jiao Tong University. His research interests are machine learning and vision. He serves as Area Chair for AAAI22, CVPR21, ACM-MM21, ICPR20.
\end{IEEEbiography}

% You can push biographies down or up by placing
% a \vfill before or after them. The appropriate
% use of \vfill depends on what kind of text is
% on the last page and whether or not the columns
% are being equalized.

%\vfill

% Can be used to pull up biographies so that the bottom of the last one
% is flush with the other column.
%\enlargethispage{-5in}

% that's all folks
\end{document}